\documentclass[mnsc]{informs4}

\OneAndAHalfSpacedXI 

\usepackage{mathrsfs}
\usepackage{lmodern}
\usepackage{fix-cm}

\usepackage{endnotes}
\let\footnote=\endnote
\let\enotesize=\normalsize
\def\notesname{Endnotes}%
\def\makeenmark{$^{\theenmark}$}
\def\enoteformat{\rightskip0pt\leftskip0pt\parindent=1.75em
	\leavevmode\llap{\theenmark.\enskip}}

\usepackage{graphicx}
\usepackage{multirow}
\usepackage{hhline}
\usepackage{booktabs}
\usepackage{comment}
\usepackage[small, margin=1cm]{caption}
\usepackage[title]{appendix}

\usepackage{color}
\definecolor{strcolor}{rgb}{0.6, 0.2, 0.6}
\definecolor{commentcolor}{rgb}{0.3125, 0.5, 0.3125}
\definecolor{keycol}{rgb}{0, 0, 1}
\usepackage{array}
\usepackage{ltablex}
\keepXColumns

\usepackage{bbm}

\usepackage{listings}
\usepackage{algorithm}
\usepackage{algorithmic}
\usepackage{hyperref}
\usepackage{url}
\usepackage{subfiles} 
\usepackage{makecell}

\newcommand {\bea}{\begin{eqnarray}}
	\newcommand {\eea}{\end{eqnarray}}

\def\blot{\quad \mbox{$\vcenter{ \vbox{ \hrule height.4pt
				\hbox{\vrule width.4pt height.9ex \kern.9ex \vrule width.4pt}
				\hrule height.4pt}}$}}

\usepackage{natbib}
\bibpunct[, ]{(}{)}{,}{a}{}{,}%
\def\bibfont{\fontsize{8}{9.5}\selectfont}%
\TheoremsNumberedThrough     
\ECRepeatTheorems

\EquationsNumberedThrough   

\gdef\AQ#1{}
\gdef\CQ#1{}

\usepackage{graphicx}
\usepackage{subcaption}
\emergencystretch=\maxdimen
\makeatletter
\newcommand{\otherlabel}[2]{\protected@edef\@currentlabel{#2}\label{#1}}
\makeatother
\usepackage{dsfont}
\usepackage{ulem}
\usepackage{tablefootnote}
\usepackage{threeparttable}
\usepackage[resetlabels]{multibib}
\newcites{ec}{References} 
\usepackage{blindtext}
\usepackage{lastpage}
\usepackage{fancyhdr}

\usepackage{longtable}
\usepackage{array}
\newcolumntype{P}[1]{>{\raggedright\arraybackslash}p{#1}}

\usepackage{enumitem}

\newcites{de}{References}
\newcites{ae}{References}
\newcites{rone}{References}
\newcites{rtwo}{References}
\newcites{rthree}{References}

\begin{document}
	
\def\COPYRIGHTHOLDER{INFORMS}%
\def\COPYRIGHTYEAR{2017}%
\def\DOI{\fontsize{7.5}{9.5}\selectfont\sf\bfseries\noindent https://doi.org/10.1287/opre.2017.1714\CQ{Word count = 9740}}

	\RUNAUTHOR{Mao et al.} %

	\RUNTITLE{Early Detection of Misinformation for Infodemic Management}

\TITLE{Early Detection of Misinformation for Infodemic Management: A Domain Adaptation Approach }


	\ARTICLEAUTHORS{

\AUTHOR{Minjia Mao}
\AFF{Lerner College of Business and Economics,
University of Delaware, Newark, DE, USA}

\AUTHOR{Xiaohang Zhao*}
\AFF{School of Information Management \& Engineering, Shanghai University of Finance and Economics, Shanghai, China}

\AUTHOR{Xiao Fang*}
\AFF{Lerner College of Business and Economics,
University of Delaware, Newark, DE, USA}

\small *Corresponding author


}
	

\ABSTRACT{
An infodemic refers to an enormous amount of true information and misinformation disseminated during a disease outbreak. Detecting misinformation at the early stage of an infodemic is key to reduce its harm to public health. An early stage infodemic is characterized by a large volume of unlabeled information concerning a disease. As a result, conventional misinformation detection methods are not suitable for this misinformation detection task because they rely on labeled information in the infodemic domain to train their models. To address this limitation, state-of-the-art methods learn their models using labeled information in other domains to detect misinformation in the infodemic domain. The efficacy of these methods depends on their ability to mitigate both covariate shift (i.e., differences in feature distributions) and concept shift (i.e., differences in labeling patterns) between the infodemic domain and the domains from which they leverage labeled information. However, these methods focus on mitigating covariate shift but overlook concept shift, rendering them less effective for the task. In response, we theoretically show the necessity of tackling both covariate and concept shifts as well as how to operationalize each of them. Built on the theoretical analysis, we develop a novel misinformation detection method that addresses both covariate and concept shifts. Using real-world datasets, we conduct extensive empirical evaluations to demonstrate the superior performance of our method over state-of-the-art misinformation detection methods as well as prevalent domain adaptation methods that can be tailored to solve the misinformation detection task. 
}






	
	%
	
\maketitle

\section{Introduction } 
\label{sec:intro}

An infodemic refers to an overwhelming volume of true and false information spread during a disease outbreak \citep{van2022misinformation}. 
Misinformation, or false information, in an infodemic misleads the public about the disease and causes significant harm to public health \citep{imhoff2020bioweapon,freeman2022coronavirus,van2022misinformation}. For example, the misinformation that the Ebola virus is intentionally created by the Congolese government to eliminate people in the city of Beni has led to attacks on Ebola clinics by local residents, hindering timely treatments for Ebola.\footnotemark\footnotetext{See \url{https://news.un.org/en/story/2019/03/1034381} (last accessed on May 28, 2024)} 
More recently, during the outbreak of the coronavirus disease (COVID-19), a significant amount of misinformation has been diffused on social media. Examples include the misinformation that baking soda can cure coronavirus,\footnotemark\footnotetext{See \url{https://apnews.com/article/archive-fact-checking-8736262219} (last accessed on May 28, 2024)} the false claim that injecting disinfectant can prevent the virus \citep{borah2022injecting}, and the incorrect attribution of the outbreak in Italy to Middle East illegal immigrants.\footnotemark\footnotetext{See \url{https://time.com/5789666/italy-coronavirus-far-right-salvini/} (last accessed on May 28, 2024)}
Such widespread of misinformation about COVID-19 has caused social panic, led people to dismiss health guidance, and weakened their confidence in vaccines, ultimately undermining pandemic response efforts \citep{bursztyn2020misinformation}.
Therefore, it is imperative to manage an infodemic to mitigate its adverse impact on public health \citep{freeman2022coronavirus,hwang2025nudge}.

According to the World Health Organization (WHO), infodemic management is the use of risk- and evidence-based approaches to manage an infodemic and reduce its harm to public health.\footnotemark\footnotetext{See \url{https://www.who.int/health-topics/infodemic\#tab=tab_1} (last accessed on May 28, 2024)} The key to effective infodemic management is the detection of misinformation at the early stage of an infodemic \citep{buchanan2020people,van2022misinformation}. Early identification of misinformation discourages people from sharing it and prevents it from reaching a much larger population, thereby mitigating its potential harm to public health \citep{buchanan2020people}. 
Moreover, the likelihood that a person trusts misinformation increases as the person is exposed longer to the misinformation \citep{zajonc2001mere,moravec2019fake}.
Early identification of misinformation reduces individuals' exposure time to misinformation, thereby lowering their likelihoods of trusting it. This, in turn, increases their chances of adhering to health guidance, ultimately benefiting public health as a whole.

An early stage infodemic features two salient characteristics. First, an emerging health event triggers a huge amount of true and false information spreading on various media platforms in a short time period. 
For example, during the early stage of the COVID-19 infodemic, a significant volume of true and false information concerning COVID-19 is diffused on social media \citep{yue2022contrastive}, and the WHO declares a worldwide infodemic \citep{van2022misinformation}. 
Second, it requires expert knowledge to distinguish between true information and misinformation in an infodemic. Moreover, during the early stage of a disease outbreak and its ensuing infodemic, even experts have no or limited knowledge about the disease, making it even more difficult to discern true information from misinformation. For example, during the early stage of the Ebola outbreak, scientists lack knowledge about the disease and its treatment \citep{adebimpe2015relevance}. Consequently, it is common that information disseminated at the early stage of an infodemic is unlabeled. As there is no labeled information (i.e., true or false), it is impossible to learn a misinformation detection model using conventional misinformation detection methods such as \cite{abbasi2010detecting}, \cite{nan2021mdfend}, and \cite{wei2022combining}, which require labeled information in their training data.


A viable solution is to leverage labeled information in other domains to build a misinformation detection model for the infodemic domain. Indeed, popular fact-checking websites like PolitiFact\footnotemark\footnotetext{See \url{https://www.politifact.com/} (last accessed on May 28, 2024)} and Gossipcop\footnotemark\footnotetext{See \url{https://en.wikipedia.org/wiki/Gossip_Cop} (last accessed on May 28, 2024)} contain a wealth of verified true and false information in domains such as politics and entertainment.
Moreover, misinformation across domains shares some common characteristics. For example, compared to true information, misinformation spreads faster, farther, and more widely \citep{oh2013community,vosoughi2018spread}. As another example, fake news, an important type of misinformation, often features emotional headlines and demonstrates inconsistency between their headlines and contents \citep{siering2016detecting}. Because of these common characteristics shared across domains, it is possible to learn a misinformation detection model using labeled information in other domains and transfer it to the infodemic domain. However, information in each domain also has its unique characteristics. Understandably, the choice of words and the organization of words to convey sentiments and semantics differ across domains. For example, less than 20\% of words used in politics news overlaps with those found in entertainment news \citep{shu2020fakenewsnet}. 
Because each domain possesses unique characteristics, a misinformation detection model learned with labeled information in other domains usually performs poorly in the infodemic domain.

\begin{figure}[htbp]
  \centering
    \caption{The Presence of Distribution Shifts across Domains.}
      \label{fig:body:shift}
  \begin{subfigure}[b]{0.8\textwidth}
    \includegraphics[width=\textwidth]{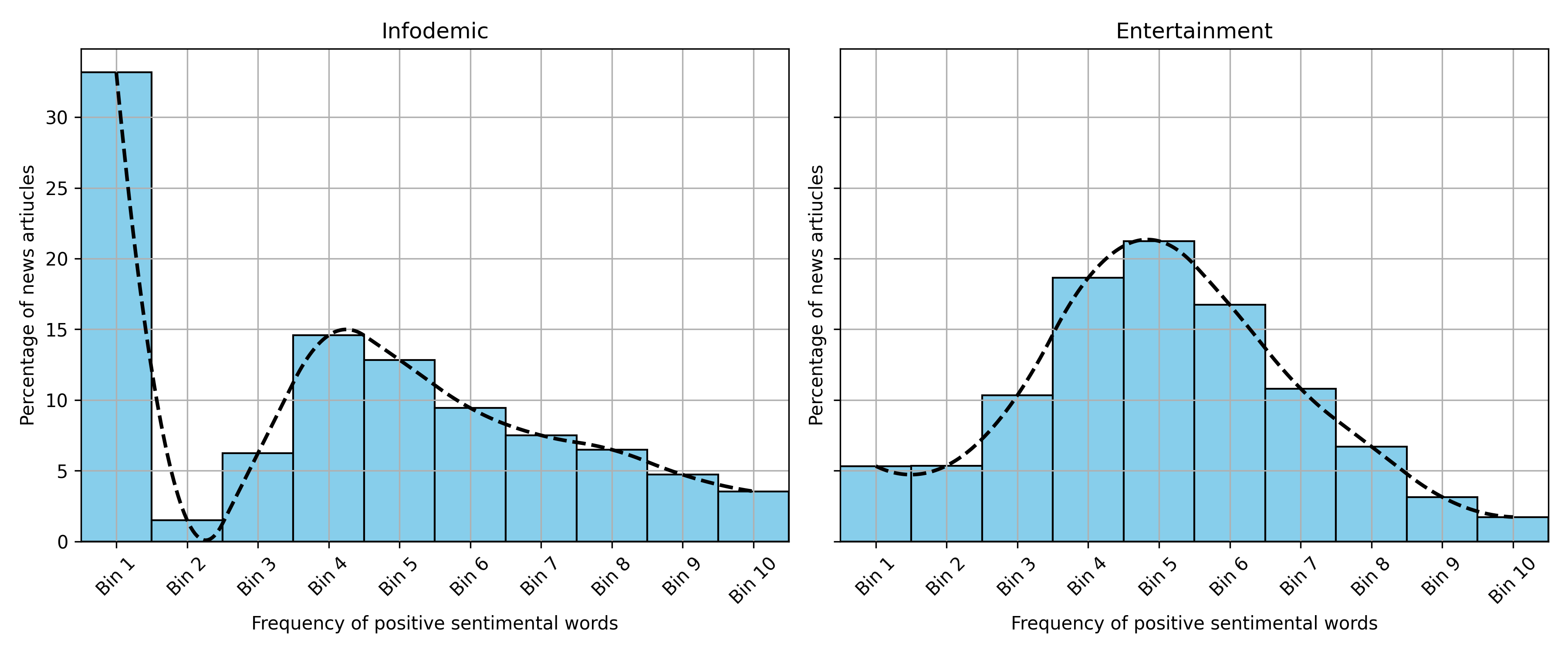}
    \captionsetup{font=small}
    \caption{Covariate Shift between Infodemic and Entertainment domains.}
    \label{fig:body:covariate}
  \end{subfigure}
  \begin{subfigure}[b]{0.8\textwidth}
    \includegraphics[width=\textwidth]{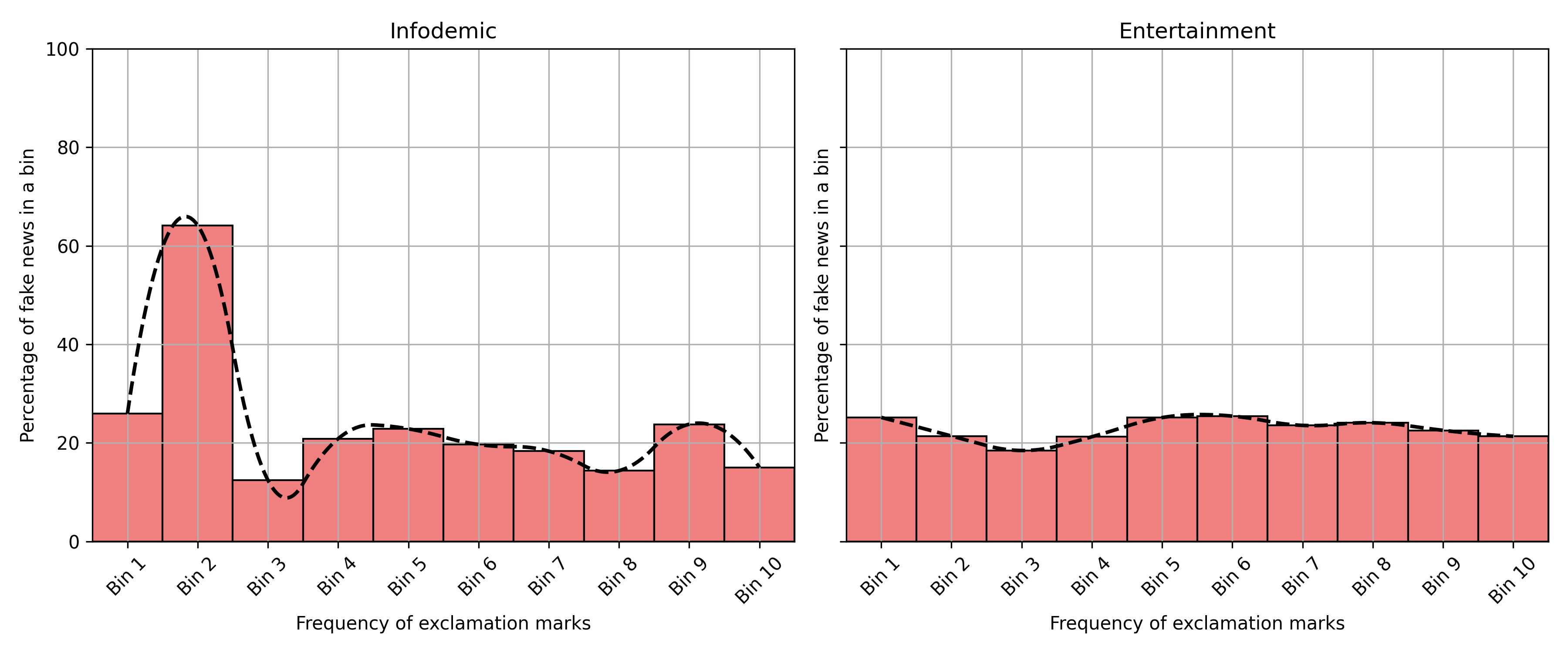}
    \captionsetup{font=small}\caption{Concept Shift between Infodemic and Entertainment domains.}
    \captionsetup{justification=raggedright,singlelinecheck=false,font=small}
    \caption*{\hspace*{0pt}\textit{Note.} Better view in color.}
    \label{fig:body:concept}
  \end{subfigure}
\end{figure}

Domain adaptation is a widely employed technique to enhance the performance of a model trained in one domain but applied to another \citep{ben2010theory}. It aims to mitigate disparities between data in different domains such that a model learned using data in one domain performs well in another domain \citep{ben2010theory}. Concretely, data in a domain is represented as a joint distribution $p(\textbf{x},y),\textbf{x}\in\mathcal{X},y\in\mathcal{Y}$, where $\mathcal{X}$ denotes the feature space and $\mathcal{Y}$ is the label space \citep{liu2022deep}. In the context of misinformation detection, $\textbf{x}$ might be explicit features, such as the number of words in a piece of information, or the embedding of a piece of information \citep{zhou2020survey}. Label $y$ denotes whether a piece of information is true or false. The joint distribution $p(\textbf{x},y)$ can be expressed as the product of $p(\textbf{x})$ and $p(y|\textbf{x})$, i.e., $p(\textbf{x},y)=p(\textbf{x})p(y|\textbf{x})$. 
Accordingly, to effectively alleviate disparities between data in different domains, it is essential to reduce both their differences in terms of $p(\textbf{x})$ and their differences in terms of $p(y|\textbf{x})$, where the former type of differences is referred to as \textit{covariate shift} and the latter type of differences is known as \textit{concept shift} \citep{liu2022deep}. 
Figure \ref{fig:body:shift} illustrates the presence of covariate shift and concept shift. 
For covariate shift, Figure \ref{fig:body:covariate} shows that the distribution of a feature (i.e., the frequency of positive sentimental words in a news article) is skewed toward bin 1 in the infodemic domain, while it exhibits a bell-shaped curve in the entertainment domain.
Figure \ref{fig:body:concept} demonstrates concept shift by showing that the percentage of fake news conditioned on a feature (i.e., the frequency of exclamation marks in a news article) differs markedly between the two domains. Specifically, in the infodemic domain, the percentage of fake news spikes in bin 2, while it is stable across bins in the entertainment domain. A more detailed explanation of the figure is provided in Appendix \ref{appendix:dist_shift}.

Existing domain adaptation methods focus on mitigating covariate shift between domains \citep{ganin2016domain, long2017deep, zhu2019aligning, peng2019moment}.
By applying these domain adaptation methods, \cite {huang2021dafd,li2021multi,yue2022contrastive} develop state-of-the-art misinformation detection methods, which learn a model from labeled information in one or more source domains to identify misinformation in a target domain with no labeled information. Although these misinformation detection methods are applicable to our problem, their efficacy is limited because they fail to account for concept shift.
In response to the research gaps, this study contributes to the literature with a novel misinformation detection method that tackles both covariate shift and concept shift. Our method design is grounded in our theoretical analysis, which shows the necessity of addressing both covariate shift and concept shift for effective domain adaptation as well as how to operationalize each of them. Guided by the theoretical analysis, we design two modules that respectively mitigate covariate shift and concept shift.
Therefore, the primary methodological difference between our method and state-of-the-art domain adaptation-based misinformation detection methods (e.g., \cite{yue2022contrastive}) lies in its module that mitigates concept shift. This module is directly informed by the theoretical analysis and provides a concrete way to operationalize and mitigate concept shift. Our method also differs from conventional misinformation detection methods such as \cite{wei2022combining}. Specifically, our method is a cross-domain method that learns a model from labeled information in other domains to classify unlabeled information in the infodemic domain. In contrast, conventional misinformation detection methods are single-domain methods that learn a model from labeled information in a domain (e.g., infodemic domain) to classify unlabeled information in the same domain.

\section{Related Work }
\label{sec:rw}

In this section, we review the Information Systems (IS) literature on misinformation, which not only situates our study in the IS literature but also reveals research gaps that motivate our study. 
We also survey existing methods for misinformation detection, which can be classified into three categories: single-domain, multi-domain, and cross-domain approaches. We then highlight the methodological novelty of our study.

\subsection{IS Literature on Misinformation and Intervention Strategies}
\label{sec:IS}

The IS literature on misinformation can be grouped into two themes: understanding why users believe and spread misinformation on social media platforms, and examining the effectiveness of intervention strategies mitigating such beliefs and behaviors. The belief of misinformation is driven by complex cognitive and contextual factors. \cite{moravec2019fake} provide neurocognitive evidence that users tend to believe news headlines aligning with their political views, even when these headlines are labeled as misinformation. Their findings suggest a strong confirmation bias, in which ideological concordance takes precedence over analytical judgment. 
\cite{pennycook2021psychology} challenge the prevailing notion that ideological motivation is the dominant driver of believing and sharing misinformation, instead suggesting that misinformation sharing is attributed to simple inattention. 
Adding another dimension, \cite{mostagir_learning_2022} study social learning in the presence of misinformation, showing that Bayesian agents equipped with sophisticated reasoning capabilities can be systematically wrong about the truth in environments with a large amount of misinformation. 

Information content plays a key role in influencing the spread of misinformation. \cite{london_seems_2022} show that people share unverified messages, such as rumors, based on both content plausibility and non-content factors like vividness and sender credibility.
\cite{seol_lying_2024} examine the phenomenon of commingled partisan falsehoods --- false statements embedded within otherwise accurate news articles. Through field and experimental studies, they show that such hybrid content significantly increases user sharing, especially when the content is distributed by reputable media firms. 
Spreading misinformation is also driven by broader social contextual factors. \cite{oh2013community} show that source ambiguity and increased public anxiety are key drivers of rumor propagation on social media during crisis events. 
\cite{mostagir_social_2023} use a network model to examine how structural inequality in access to accurate information affects vulnerability to misinformation. Their findings show that communities with limited access to knowledgeable agents are more susceptible to manipulation, and that moderate levels of inequality can, in some cases, result in worse misinformation outcomes than extreme inequality. 

The effectiveness of intervention strategies  has also been examined. One strategy is the use of fact-checking labels. \cite{moravec_appealing_2020} test both the intuitive (System 1) and deliberative (System 2) formats of such interventions and find that a combined approach is most effective in reducing the belief in misinformation. 
\cite{bhattacherjee_can_2023} discovers the subtle effects of fact checking on a user's belief in misinformation, demonstrating that its effectiveness depends on factors such as the credibility of the fact checker and the strength of the user's attitude.
Beyond tagging strategies, interventions that target users’ cognitive processes have also been investigated. \cite{moravec_you_2022} demonstrate that prompting users to rate the truthfulness of news articles activates reflective thinking and reduces their belief in fake content. \cite{hwang2025nudge} evaluate a platform-level policy adopted by Twitter, which nudges users searching for high-risk topics to credible sources. Their study shows that this policy significantly curbs the spread of misinformation by reducing both the original posting and subsequent sharing of false content. While the IS literature primarily investigates the misinformation problem from a behavioral perspective, few IS studies develop misinformation detection methods. 
In the following subsections, we review existing misinformation detection methods from both the IS and machine learning literature.  

\subsection{Single and Multi-domain Misinformation Detection Methods} \label{sec:rw_single}

Single-domain misinformation detection methods train a classification model using labeled information from a specific domain (e.g., true and fake news from the domain of politics), and then employ the model to classify unlabelled information in the same domain (e.g., unlabelled political news). In this category, knowledge-based methods classify unlabelled information by comparing the content of the information (e.g., textual content of news) against known facts \citep{ciampaglia2015computational,lee_explainable_2024}. For example, \cite{ciampaglia2015computational} retrieve pertinent facts from authoritative websites like Wikipedia, represent extracted facts as a knowledge graph, and employ the knowledge graph to automatically detect misinformation. \cite{lee_explainable_2024} build a word network of a claim and its evidence, measure the level of consistency between them based on the network, and use the consistency score to predict the claim’s veracity. Additionally, writing styles identified from the content of information, such as readability, sensationalism, informality, and subjectivity, have been utilized to detect misinformation \citep{zhou2020survey}. 
Recent methods consider the propagation of information on social media, in addition to the content of information \citep{papanastasiou2020fake}. 
Some methods employ propagation patterns to detect misinformation because misinformation spreads faster, farther, and more widely compared to true information \citep{vosoughi2018spread}. Others harness the collective intelligence of crowds, such as likes, shares, and comments from social media users \citep{atanasov2017distilling,pennycook2021psychology}. For example, \cite{wei2022combining} employ both the content of news articles and social media users' comments on these news articles to detect fake news.






Multi-domain misinformation detection methods learn a classification model from labeled information in various domains (e.g., true and fake news in the politics, health, and sports domains), and then employ the model to classify unlabelled information in each of these domains \citep{nan2021mdfend,zhu2022memory}. 
The rationale underlying these methods is to leverage labeled information in multiple domains to enhance misinformation detection in each individual domain.
However, because of variations in information content among different domains, a model learned with labeled information in these domains exhibits significant performance differences when applied to each of these domains \citep{zhu2022memory}. That is, the effectiveness of the learned model in identifying misinformation in one domain could be significantly inferior to its effectiveness in detecting misinformation in another domain. To tackle this problem, several methods incorporate domain-specific characteristics into the model training. For example, 
\cite{nan2021mdfend} propose to embed different domains as different vectors and integrate these vectors into the learning of a multi-domain misinformation detection model. \cite{zhu2022memory} introduce a domain adapter for each specific domain, where each adapter is a collection of representative document embeddings from the corresponding domain. Representations of news articles in each domain are then adjusted by their corresponding domain adapter to train a multi-domain model. 

\subsection{Cross-domain Misinformation Detection Methods} \label{sec:rw:cross}

Cross-domain misinformation detection methods leverage labeled information in one or more source domains (e.g., true and fake news in the politics and sports domains) to classify unlabelled information in a target domain (e.g., unlabelled news in the health domain). Unlike multi-domain methods, which identify misinformation in various domains, cross-domain methods focus on detecting misinformation in one target domain. Depending on whether training data contain labeled or unlabeled information in a target domain, cross-domain methods can be further categorized into two subgroups.  

One subgroup of cross-domain methods learns a model from training data consisting of labeled information in source domains as well as labeled information in a target domain to classify unlabeled information in the target domain \citep{mosallanezhad2022domain,nan2022improving}.
These methods essentially utilize labeled source domain information as out-of-distribution data to improve the performance of misinformation detection in the target domain \citep{zhang2023domain}. For example, \cite{nan2022improving} pretrain a misinformation detection model using labeled information in source and target domains. Next, a language model is learned from labeled target domain information. Each piece of source domain information is then assigned a transferability score based on the degree to which the language model can predict its content. Finally, the pretrained model is fine-tuned using labeled source domain information weighted by their corresponding transferability scores as well as labeled target domain information.

The other subgroup learns a model from training data consisting of labeled information in source domains and unlabeled information in a target domain. Since there is no labeled target domain information, this subgroup of methods aims to transfer a model learned from labeled source domain information to the target domain. To this end, domain adaptation is a suitable and widely adopted technique. In the following, we review domain adaptation methods and their applications in cross-domain misinformation detection. 
In the literature of domain adaption, a domain consists of a feature space $\mathcal{X}$, a label space $\mathcal{Y}$, and a probability distribution $p(\textbf{x},y), \textbf{x} \in \mathcal{X}$ and $y \in \mathcal{Y}$ \citep{liu2022deep}. As the probability distribution in a source domain usually differs from that in a target domain, a model learned from source domain data often performs poorly in the target domain \citep{ben2010theory}. To tackle this issue, it is necessary to reduce the disparity between source and target domains in terms of $p(\textbf{x},y)$. Existing domain adaptation methods concentrate on mitigating the difference in feature distribution (i.e., $p(\textbf{x})$) between source and target domains, known as covariate shift \citep{kouw2018introduction}.\footnotemark\footnotetext{In the field of data stream mining, concept shift or concept drift refers to the phenomenon that data samples with same features could have different labels at different timestamps \citep{roychowdhury2023tackling}. Given a model trained with previous data samples and a continuous influx of new data samples, the objective of data stream mining methods is to determine when and how to retrain the model \citep{vorburger2006entropy, fang2013right,roychowdhury2023tackling}. These methods are not applicable to our problem because they require labeled data samples at all timestamps while instances of target domain information are unlabeled in our problem. Moreover, these methods aim to decide when and how to retrain a model in a dynamic environment, whereas the objective of our problem is to classify unlabeled target domain instances.}
In this vein, \cite{ganin2016domain} propose a domain adversarial network to learn domain-invariant features from data in source and target domains. \cite{saito2018maximum} devise a mini-max mechanism to align feature distributions between source and target domains. In the maximum stage, two classifiers are learned from source samples such that the discrepancy between their classifications of target samples is maximized. In the minimum stage, features of target samples are adjusted to minimize such discrepancy. 
\cite{rangwani2022closer} develop a domain adaption method that incorporates a smoothing mechanism into the method proposed by \cite{ganin2016domain}.
\cite{rostami2023overcoming} construct a pseudo-dataset using a model learned from labeled source domain data and then mitigate feature distribution difference between the pseudo-dataset and target domain data.\footnotemark\footnotetext{Although the title of the paper contains the phrase of concept shift, it actually mitigates covariate shift. This is evident in the objective function of the proposed algorithm, i.e., Equation (3) in \citep{rostami2023overcoming}. According to the equation, the algorithm aims to mitigate the difference between the feature distribution $p_T(\textbf{X}_T)$ of target domain data and the feature distribution $\hat{p}_J(\textbf{Z}_p)$ of the pseudo-dataset.}
While the methods reviewed above implicitly minimize covariate shift between source and target domains, there are methods explicitly measure and minimize covariate shift. For example, \cite{long2015learning} utilize a metric known as the multi-kernel Maximum Mean Discrepancy (MMD) to quantify covariate shift, integrate this metric into a loss function, and minimize covariate shift by minimizing the loss function. In a follow-up study, \cite{long2017deep} introduce another metric , known as the joint MMD, to measure covariate shift. In addition, \cite{chen2022reusing} employ the nuclear-norm Wasserstein discrepancy to evaluate covariate shift between source and target domains.
Recent studies have employed domain adaptation techniques to solve the cross-domain misinformation detection problem with unlabeled target domain information. 
For example, \cite{li2021multi} extend the work of \cite{ganin2016domain} to detect misinformation in a target domain by leveraging labeled information in multiple source domains whereas \cite{ng2023augmenting} apply the domain adversarial network developed by \citep{ganin2016domain} to identify fake news and reviews. \cite{huang2021dafd} utilize the MMD metric 
to measure covariate shift between source domain information and target domain information, and mitigate covariate shift by minimizing the metric. 
\cite{yue2022contrastive} develop a variant of the MMD metric to measure covariate shift between source and target domain information.

\begin{table*}[htbp]
\centering
\caption{Comparison Between Our Method and Existing Misinformation Detection Methods. }
\begin{tabular}{p{6.8cm} p{2.5cm}<{\centering} p{2.5cm}<{\centering} p{2.5cm}<{\centering}}
    \toprule 
    & Applicable to our problem & Addressing covariate shift & Addressing concept shift \\
    \midrule 
    Single- and multi-domain misinformation detection methods, e.g., \cite{wei2022combining}, \cite{zhu2022memory}  & No & No & No \\

    \specialrule{0em}{2pt}{2pt}

    Cross-domain misinformation detection methods with labeled target domain information, e.g., \cite{mosallanezhad2022domain}, \cite{nan2022improving} & No & Yes & No \\
    
    \specialrule{0em}{2pt}{2pt} 

    Cross-domain misinformation detection methods with unlabeled target domain information, e.g., \cite{li2021multi}, \cite{yue2022contrastive} & Yes & Yes & No \\
    
    \specialrule{0em}{2pt}{2pt}
    
    Our method & Yes & Yes & Yes \\

    \bottomrule
\end{tabular}
\label{tab:rw}
\end{table*}%

\subsection{Key Novelty of Our Study }
\label{sec:novelty}

Our literature review suggests the following research gaps, as summarized in Table \ref{tab:rw}. First, existing single and multi-domain misinformation detection methods as well as cross-domain methods with labeled target domain information are not applicable to our problem. These methods require labeled target domain information (i.e., the infodemic domain in our study) in their training datasets, while our problem involves completely unlabeled information in the infodemic domain. Second, although existing cross-domain methods with unlabeled target domain information are applicable to our problem, they are less effective in solving the problem because they fail to address concept shift. In fact, none of the existing misinformation detection methods is capable of tackling concept shift. However, to ensure the efficacy of a model learned using labeled source domain information in classifying unlabeled target domain information, it is essential to mitigate both covariate shift and concept shift between source and target domains \citep{liu2022deep}. To this end, we propose a misinformation detection method that tackles both covariate shift and concept shift, in contrast to existing methods that focus on covariate shift solely. Therefore, the key methodological novelty of our method lies in its addressing of concept shift. To implement this methodological novelty, our study features two innovations. First, our theoretical analysis of domain adaptation not only underscores the importance of addressing both covariate shift and concept shift but also shows how to operationalize each of them. Second, informed by our theoretical analysis, we introduce a novel misinformation detection method, two modules of which are respectively designed to tackle covariate shift and concept shift. 
In Appendix \ref{appendix:compare}, we further demonstrate the practical advantages of our method by comparing it with existing misinformation detection methods along additional dimensions beyond domain adaptation.

\section{Problem Formulation } \label{sec:problem_formulation}

We consider a misinformation detection problem at the early stage of an infodemic, when all instances of information in this domain are unlabeled. Concretely, let $\mathcal{D}_T$ be a dataset of $N_T$ pieces of unlabeled information in the infodemic domain. The subscript $T$ signifies target domain, which is the infodemic domain in this study. As an example, a piece of unlabeled information might encompass the textual content of a news article about COVID-19.
Let $\textbf{x}_i^T$ denote a set of features extracted from a piece of unlabeled information in $\mathcal{D}_T$, $i=1,2,\dots,N_T$. We aim to predict the label (true or false) for each piece of information in $\mathcal{D}_T$. 

To accomplish this objective, we are given $k$ datasets of labeled information, i.e., $\mathcal{D}_{S_1}, \mathcal{D}_{S_2},\dots, \mathcal{D}_{S_k}$. The subscript $S$ refers to source domain, which provides labeled information to facilitate the label prediction in the target domain. Concretely, $\mathcal{D}_{S_j}$ consists of $N_{S_j}$ pieces of labeled information, $j=1,2,\dots,k$. Considering an example of $S_j$ being the politics domain, a piece of information in this domain could comprise the textual content of a political news article.
Each piece of labeled information in $\mathcal{D}_{S_j}$ is represented as $(\textbf{x}_i^{S_j},y_i^{S_j})$, where $\textbf{x}_i^{S_j}$ denotes features extracted from the information, $y_i^{S_j}$ is the label of the information, and $i=1,2,\dots,N_{S_j}$.  Label $y_i^{S_j}=0$ indicates that the information is true and $y_i^{S_j}=1$ shows that the information is false.
We now formally define the problem of early detection of misinformation for infodemic management.

\textbf{Definition 1 (Early Detection of Misinformation Problem (EDM)).} 
Given a dataset $\mathcal{D}_T$ of $N_T$ pieces of unlabeled information in the infodemic domain and $k$ datasets of labeled information in various source domains, $\mathcal{D}_{S_1}, \mathcal{D}_{S_2},\dots, \mathcal{D}_{S_k}$, where $\mathcal{D}_{S_j}$ consists of $N_{S_j}$ pieces of labeled information, $j=1,2,\dots,k$, the objective of the problem is to learn a model from the data to classify each piece of information in $\mathcal{D}_T$ as true or false.

\section{Method } \label{sec:4}

In this section, we theoretically analyze the performance of a model learned using labeled source domain instances but applied to classify unlabeled target domain instances. Guided by the insights from our theoretical analysis, we then propose a novel method to solve the EDM problem.

\subsection{Theoretical Analysis} \label{sec:4.1}

We start with the setting of one target domain $T$ and one source domain $S$. Let $\mathcal{D}_T$ be a dataset of $N_T$ unlabelled instances in $T$ and $\mathcal{D}_S$ be a dataset of $N_S$ labeled instances in $S$. In the context of misinformation detection, an instance is a piece of information.  
We denote $f_T$ 
as the true labeling function in the target domain that maps an instance with features $\textbf{x}$ into the probability of the instance being 1, i.e., $f_T(\textbf{x})=p_T(y=1|\textbf{x})$. Similarly, $f_S$
is the true labeling function in the source domain and $f_S(\textbf{x})=p_S(y=1|\textbf{x})$. Let $h$
be a hypothesis learned from labeled instances in $\mathcal{D}_S$ and $h(\textbf{x})=p_h(y=1|\textbf{x})$. Following \cite{ben2010theory}, we define the source error $\epsilon_{S}(h)$ of hypothesis $h$ as the expected difference between $h$ and the true labeling function $f_S$ in the source domain
\begin{equation} \label{eq:srcerror}
    \epsilon_{S}(h) = E_{\textbf{x}\sim \mathbb{D}_S} 
    [|h(\textbf{x})-f_S(\textbf{x})| ],
\end{equation}  
where $\mathbb{D}_S$ represents the feature distribution in the source domain. In a similar manner, we define the target error $\epsilon_{T}(h)$ of hypothesis $h$ as the expected difference between $h$ and the true labelling function $f_T$ in the target domain
\begin{equation}
    \epsilon_{T}(h) = E_{\textbf{x}\sim \mathbb{D}_T} [|h(\textbf{x})-f_T(\textbf{x})| ],
\end{equation}
where $\mathbb{D}_T$ denotes the feature distribution in the target domain. Recall that domain adaptation aims to learn $h$ from labeled instances in $\mathcal{D}_S$ to classify each unlabelled instance in $\mathcal{D}_T$. Thus, the objective of domain adaptation is to minimize the target error $\epsilon_{T}(h)$. 
To achieve this objective, we show a bound of $\epsilon_{T}(h)$ in the following theorem.


\textbf{Theorem 1} 
\textit{For a source domain instance with features $\textbf{x}_i^S$, let $c(\textbf{x}_i^{S})$ be the features of its nearest target domain instance, where $i=1,2,\dots,N_S$ and the distance between a pair of instances are measured over their feature space. If $f_T$ is $L$-Lipschitz continuous,\footnotemark\footnotetext{The L-Lipschitz continuous assumption is commonly used in theoretical analyses of machine learning algorithms \citep[e.g.,][]{arjovsky2017wasserstein,asadi2018lipschitz,kim2021lipschitz}. } then for any $\eta\in (0,1)$, with probability at least $(1-\eta)^2$, } 
\begin{equation} \label{eq:theorem1}
\begin{aligned}
\epsilon_T(h) \leq \epsilon_S(h) + \hat{d}_{\mathcal{H}}(\mathcal{D}_S,\mathcal{D}_T) + \frac{1}{N_S} \sum_{i=1}^{N_S} |f_S(\textbf{x}_i^S) - f_T(c(\textbf{x}_i^{S}))| + \frac{L}{N_S} \sum_{i=1}^{N_S} ||\textbf{x}_i^S - c(\textbf{x}_i^{S})||  + C_1, 
\end{aligned}
\end{equation} 
\textit{where $L$ is the Lipschitz constraint constant, $C_1$ is a constant, $\hat{d}_{\mathcal{H}}(\mathcal{D}_S,\mathcal{D}_T)$ denotes the empirical $\mathcal{H}$-divergence between distributions $\mathbb{D}_S$ and $\mathbb{D}_T$, estimated using features of source domain instances in $\mathcal{D}_S$ and features of target domain instances in $\mathcal{D}_T$, and $||\textbf{x}_i^S - c(\textbf{x}_i^{S})||$ represents the distance between $\textbf{x}_i^S$ and $c(\textbf{x}_i^{S})$.
}
\otherlabel{theorem:single}{1}

\textit{Proof. See Appendix~\ref{appendix:proof1}. }

Theorem~\ref{theorem:single} establishes an upper bound of $\epsilon_T(h)$, which can be empirically computed using datasets $\mathcal{D}_S$ and $\mathcal{D}_T$ drawn from distributions $\mathbb{D}_S$ and $\mathbb{D}_T$, respectively. By Theorem~\ref{theorem:single}, to minimize $\epsilon_T(h)$, we need to minimize the first three terms of its upper bound specified in Inequality~(\ref{eq:theorem1}). This is because the fourth term of the upper bound is inherently minimized, given that $c(\textbf{x}_i^{S})$ is the closest to $\textbf{x}_i^{S}$, and the last term $C_1$ is a constant.

The first term $\epsilon_S(h)$ can be minimized by learning a classification model from labeled source domain instances.  
The second term $\hat{d}_{\mathcal{H}}(\mathcal{D}_S,\mathcal{D}_T)$ is the empirical estimation of the 
$\mathcal{H}$-divergence between the source domain feature distribution $\mathbb{D}_S$ and the target domain feature distribution $\mathbb{D}_T$. $\mathcal{H}$-divergence has been extensively utilized in domain adaptation studies to measure distances between distributions. Please refer to Appendix~\ref{appendix:definitions} for a description of $\mathcal{H}$-divergence and its empirical estimation $\hat{d}_{\mathcal{H}}(\mathcal{D}_S,\mathcal{D}_T)$. Fundamentally, minimizing the $\mathcal{H}$-divergence between source and target domain feature distributions is to minimize covariate shift between source and target domains. Therefore, existing domain adaptation methods for mitigating covariate shift, such as \cite{ganin2016domain} and \cite{zhu2019aligning}, can be applied to minimize $\hat{d}_{\mathcal{H}}(\mathcal{D}_S,\mathcal{D}_T)$. 



For the third term $\frac{1}{N_S} \sum_{i=1}^{N_S} |f_S(\textbf{x}_i^S) - f_T(c(\textbf{x}_i^{S}))|$, recall that  
$f_S(\textbf{x}_i^S)=p_S(y=1|\textbf{x}_i^S)$ and $f_T(c(\textbf{x}_i^{S}))=p_T(y=1|c(\textbf{x}_i^{S}))$. Hence, to minimize the third term, we need to minimize the summation of differences between 
$p_S(y=1|\textbf{x}_i^S)$ and $p_T(y=1|c(\textbf{x}_i^{S}))$, summed over all source domain instances and their corresponding nearest target domain instances. Therefore, minimizing the third term is essentially to minimize concept shift between source and target domains.




Next, we consider the setting of one target domain $T$ and multiple source domains $S_1, S_2,\dots,S_k$. Similar to the notations
used in the setting of one source domain, we denote $\mathcal{D}_T$ as a dataset of $N_T$ unlabelled instances in $T$ and $\mathcal{D}_{S_j}$ as a dataset of $N_{S_j}$ labeled instances in $S_j$, $j=1,2,\dots,k$. Let $h$ be a hypothesis learned from labeled instances in $\mathcal{D}_{S_1},\mathcal{D}_{S_2},\dots,\mathcal{D}_{S_k}$. 
We denote $f_T$ and $f_{S_j}$ as the true labeling functions in $T$ and $S_j$, respectively, $j=1,2,\dots,k$. 
For a hypothesis $h$, $\epsilon_{T}(h)$ and $\epsilon_{S_j}(h)$ represent its errors in $T$ and $S_j$, respectively, $j=1,2,\dots,k$. Built on Theorem \ref{theorem:single}, the following proposition gives an upper bound of $\epsilon_{T}(h)$ in the setting of multiple source domains.

\textbf{Proposition 1} 
\textit{For a source domain instance with features $\textbf{x}_i^{S_j}$ in $ \mathcal{D}_{S_j}$, let $c(\textbf{x}_i^{S_j})$ be the features of its nearest instance in $\mathcal{D}_T$, where $j=1,2,\dots,k$ and $i=1,2,\dots,N_{S_j}$. If $f_T$ is $L$-Lipschitz continuous, then for any $\eta\in(0,1)$, with probability at least $(1-\eta)^{2k} $, } 
\begin{equation} \label{eq:multi}
\begin{aligned}
\epsilon_T(h) \leq \frac{1}{k} \sum_{j=1}^k \{ \epsilon_{S_j}(h) + 
& \hat{d}_{\mathcal{H} }(\mathcal{D}_{S_j},\mathcal{D}_T) + \frac{1}{N_{S_j}} \sum_{i=1}^{N_{S_j}} |f_{S_j}(\textbf{x}_i^{S_j}) - f_T(c(\textbf{x}_i^{S_j}))| + \frac{L}{N_{S_j}} \sum_{i=1}^{N_{S_j}} ||\textbf{x}_i^{S_j} - c(\textbf{x}_i^{S_j})||  + C_j  \}, 
\end{aligned}
\end{equation} 
\textit{where $L$ is the Lipschitz constraint constant, $C_j$ is a constant, and $\hat{d}_{\mathcal{H}}(\mathcal{D}_{S_j},\mathcal{D}_T)$ denotes the empirical $\mathcal{H}$-divergence between the feature distribution in the source domain $S_j$ and that in the target domain $T$.}
\otherlabel{theorem:multi}{1}

\textit{Proof. See Appendix~\ref{appendix:proof1}. } 


Proposition~\ref{theorem:multi} offers useful insights for solving the EDM problem, which aims to learn a model from labeled information in multiple source domains to classify unlabeled information in the target infodemic domain with minimum error (i.e., $\epsilon_T(h)$). Informed by the proposition, we design a method that minimizes $\epsilon_T(h)$ through its three modules, each dedicated to reducing one of the first three terms in Inequality~(\ref{eq:multi}). Specifically, to reduce $\sum_{j=1}^k\epsilon_{S_j}(h)$, we implement a classification module, which is trained to accurately predict the label for each piece of source domain information. A covariate alignment module is developed to reduce $\sum_{j=1}^k\hat{d}_{\mathcal{H} }(\mathcal{D}_{S_j},\mathcal{D}_T)$, thus diminishing covariate shift between source domains and the target domain. Finally, we propose a concept alignment module to mitigate $\sum_{j=1}^k \sum_{i=1}^{N_{S_j}} |f_{S_j}(\textbf{x}_i^{S_j}) - f_T(c(\textbf{x}_i^{S_j}))|$, thereby reducing concept shift between source domains and the target domain. The concept alignment module constitutes the main methodological novelty of our method, and we designate our method as Domain Adaptation with Concept Alignment (DACA). 

\begin{figure}[htbp]
\FIGURE{\includegraphics[width=1\textwidth]{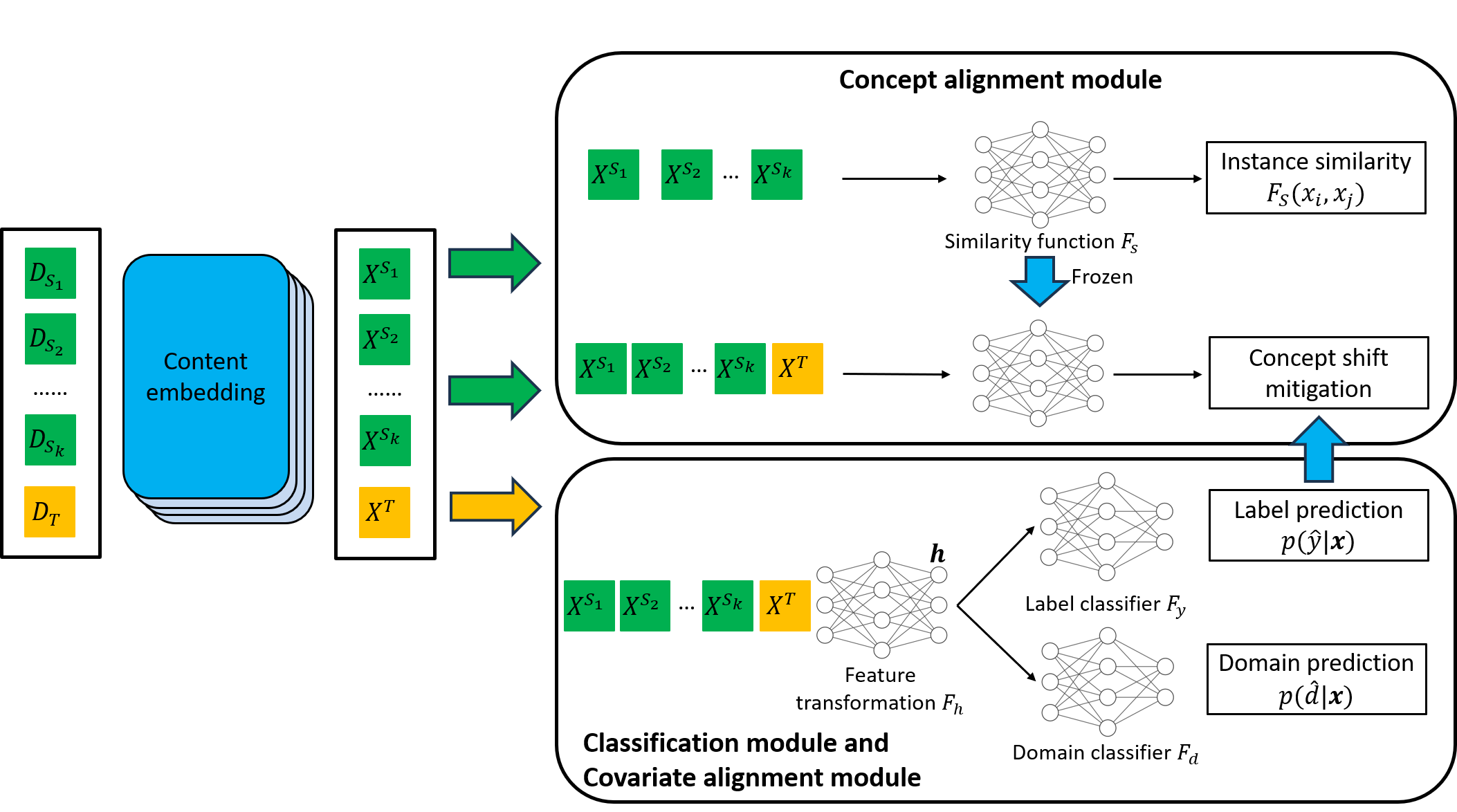}}
{Overall Architecture of Domain Adaptation with Concept Alignment (DACA) Method  \label{fig:overview}}
{Better view in color. }
\end{figure}

\subsection{Method Overview } \label{sec:4.1.2} 

Grounded in our theoretical analysis, the DACA method adeptly alleviates disparities between source and target domains through its covariate alignment and concept alignment modules. As a result, the model learned from labeled source domain information by its classification module can effectively classify unlabeled infodemic domain information. Figure~\ref{fig:overview} illustrates the overall architecture of the DACA method. 
As shown, the inputs to the method consist of $k$ datasets of labeled information in various source domains, $\mathcal{D}_{S_1}, \mathcal{D}_{S_2}, \dots, \mathcal{D}_{S_k}$, and a dataset $\mathcal{D}_{T}$ of unlabeled information in the target infodemic domain. 
A content embedding function based on the method proposed by \cite{zhu2022memory} represents each piece of input information as a vector. Specifically, it extracts and represents three types of features from an instance of input information (e.g., a news article): semantic, style, and emotion features. More concretely, the semantic feature vector is extracted by first encoding the input information as a token embedding sequence via the pretrained RoBERTa model \citep{liu2019roberta} and then aggregating the sequence into a single vector using the TextCNN model \citep{kim2014textcnn}. Style features, such as the readability score, are derived from the input information and then converted into the style feature vector via a multi-layer perceptron transformation \citep{zhu2022memory}. Similarly, emotion features, such as the counts of emoticons, pronouns, and uppercase letters, are computed from the input information and transformed into the emotion feature vector.  
Finally, the three vectors are aggregated into a single vector representing the input information through an attention mechanism \citep{zhu2022memory}.
Let $X^{S_j}=\{\textbf{x}_i^{S_j}|i=1,2,\dots,N_{S_j}\}$
and $X^{T}=\{\textbf{x}_i^T|i=1,2,\dots,N_T \}$ denote the set of embedding vectors extracted from input information in source domain $S_j$ and target domain $T$, respectively, $j=1,2,\dots,k$. These features serve as the inputs to the three modules of the DACA method. 

The design of these modules is guided by the theoretical results in Theorem \ref{theorem:single} and Proposition \ref{theorem:multi}. Specifically, to minimize the target error, it is necessary to reduce the first three terms in Inequality~(\ref{eq:multi}). Accordingly, each module is designed to reduce one of these three terms.
The classification module aims to reduce the term $\sum_{j=1}^k\epsilon_{S_j}(h)$ in Inequality~(\ref{eq:multi}). To this end, it trains a label classifier to accurately predict the label of a piece of information as true (0) or false (1). Concretely, the label classifier takes the features $\textbf{x}$ of each piece of information as input and predicts its probability of being 1, i.e., $p(\hat{y}=1|\textbf{x})$. Given that only source domain information is labeled, the label classifier is trained using labelled source domain information. 
The covariate alignment module trains a domain classifier to predict whether a piece of information is from a source domain or the target domain. As the objective of this module is to mitigate the feature distribution discrepancy between source domains and the target domain (i.e., reducing the term $\sum_{j=1}^k\hat{d}_{\mathcal{H} }(\mathcal{D}_{S_j},\mathcal{D}_T)$ in Inequality~(\ref{eq:multi})), it learns representations of information instances such that a well-trained domain classifier cannot discern whether an information instance is from a source domain or the target domain \citep{ganin2016domain}.


Finally, to mitigate the term $\sum_{j=1}^k \sum_{i=1}^{N_{S_j}} |f_{S_j}(\textbf{x}_i^{S_j}) - f_T(c(\textbf{x}_i^{S_j}))|$ in Inequality~(\ref{eq:multi}), we propose a concept alignment module.
Proposition~\ref{theorem:multi} sheds light on the design of this module.
According to the proposition, 
for each piece of information in a source domain, we need to find its nearest target domain information. Therefore, it is imperative to have a function that measures the similarity between two pieces of information over their feature space. To this end, we introduce a submodule that defines and learns a similarity function using labeled source domain information. 
Utilizing the similarity function, a concept shift mitigation submodule identifies the nearest target domain information for each piece of source domain information.\footnotemark\footnotetext{The time complexity of computing the similarity between each piece of source domain information and every piece of target domain information is $O(\sum_{j=1}^{k}N_{S_j}N_T)$, where the source domain $\mathcal{D}_{S_j}$ consists of $N_{S_j}$ pieces of labeled information, $j=1,2,\dots,k$, and the target domain $\mathcal{D}_T$ contains $N_T$ pieces of unlabeled information.}
By Proposition~\ref{theorem:multi}, this submodule then mitigates concept shift between source domains and the target domain by minimizing $\sum_{j=1}^k \sum_{i=1}^{N_{S_j}} |f_{S_j}(\textbf{x}_i^{S_j}) - f_T(c(\textbf{x}_i^{S_j}))|$,
where $f_{S_j}(\textbf{x}_i^{S_j})$ and $f_T(c(\textbf{x}_i^{S_j}))$ are estimated using the label classifier in the classification module.

\subsection{Classification Module and Covariate Alignment Module } \label{sec:lc_dc}

As depicted in Figure~\ref{fig:overview}, the classification module and the covariate alignment module share a feature transformation layer. 
This layer, denoted by $F_h$, 
takes each feature vector $\textbf{x}$ in $X^{S_1}$, $X^{S_2}$, ..., $X^{S_k}$, $X^{T}$ as input 
and outputs the transformed vector $\textbf{h}$. Formally, we have
\begin{equation} \label{eq:fh}
\begin{aligned}
    \textbf{h} = F_h(\textbf{x}) = \text{MLP}_1(\textbf{x};\theta_h),
\end{aligned}
\end{equation}
where $\text{MLP}_1$ is a multi-layer perceptron (MLP) with trainable parameters $\theta_h$.

The classification module features an MLP-based label classifier, denoted by $F_y$, which takes the transformed feature vector $\textbf{h}$ as input and outputs the probability of being labeled as 1 (i.e., false) for the piece of information characterized by feature vector $\textbf{x}$. Concretely, we have
\begin{equation}
\begin{aligned} \label{eq:cls_layer}
    p(\hat{y}=1|\textbf{x}) = F_y(\textbf{h}) = \text{Sigmoid} \big( \text{MLP}_2(\textbf{h};\theta_y) \big),
\end{aligned}
\end{equation} 
where $\hat{y}$ is the predicted label for the piece of information, $\text{MLP}_2$ is an MLP with the ReLU activation function for its internal layers and trainable parameters $\theta_y$, and $\text{Sigmoid()}$ denotes the sigmoid function that converts the output of $\text{MLP}_2$ to a probability value. 
The classification module is trained by minimizing the total cross-entropy loss summed over all labeled source domain information: 
\begin{equation} \label{eq:cls}
\begin{aligned}
    \mathcal{L}_y(\theta_h, \theta_y) &= - \sum_{\textbf{x} \in X^S } \bigg[y \log \big( p(\hat{y}=1|\textbf{x}) \big) + (1-y) \log \big( 1-p(\hat{y}=1|\textbf{x}) \big) \bigg] \\
    &= - \sum_{\textbf{x} \in X^S } \bigg[y \log F_y(F_h(\textbf{x})) + (1-y) \log \big( 1-F_y(F_h(\textbf{x})) \big) \bigg],
\end{aligned}
\end{equation}
where $X^S=\cup_{j=1}^{k} {X}^{S_j}$ 
and $y$ is the true label of $\textbf{x}$. Minimizing $\mathcal{L}_y$ trains the classification module to score high probabilities for true labels, thereby inducing a model with low source error (i.e., $\sum_{j=1}^k\epsilon_{S_j}(h)$ in Proposition~\ref{theorem:multi}).

The covariate alignment module aims to mitigate covariate shift between source domains and the target domain, i.e., reducing $\sum_{j=1}^k\hat{d}_{\mathcal{H} }(\mathcal{D}_{S_j},\mathcal{D}_T)$ in Proposition~\ref{theorem:multi}. To this end, we adapt the domain adversarial method developed by \cite{ganin2016domain} to implement this module. Let $d$ denote the true domain for a piece of information characterized by feature vector $\textbf{x}$, where $d=0$ if it belongs to a source domain and $d=1$ if it comes from the target domain. The module employs a domain classifier, denoted by $F_d$, that takes the transformed feature vector $\textbf{h}$ as input and outputs its probability of belonging to the target domain. Mathematically, we have 
\begin{equation} \label{eq:dcls_layer}
\begin{aligned}
    p(\hat{d}=1|\textbf{x}) = F_d(\textbf{h}) = \text{Sigmoid} \big( \text{MLP}_3(\textbf{h};\theta_d) \big),
\end{aligned}
\end{equation} 
where $\hat{d}$ is the predicted domain (i.e., source or target) for $\textbf{x}$, and $\text{MLP}_3$ is an MLP with the ReLU activation function for its internal layers and trainable parameters $\theta_d$. 
On the one hand, $F_d$ is trained to accurately predict the domain for each piece of information. On the other hand, to reduce covariate shift, the vector $\textbf{h}$ produced by the feature transformation layer $F_h$ 
should be domain-invariant, i.e., when presented to $F_d$, the classifier cannot tell whether it is from a source domain or the target domain 
\citep{li2021multi}. 
With this reasoning in mind, the learning of domain-invariant features can be formulated as a min-max game defined as follows\citep{ganin2016domain,li2021multi}:
\begin{equation} 
\label{eq:minmax}
\begin{aligned}    
    & \mathop{\min}_{\theta_h} \mathop{\max}_{\theta_d} \  
    \sum_{\textbf{x} \in X } \bigg[d \log \big( p(\hat{d}=1|\textbf{x}) \big) + (1-d) \log \big( 1-p(\hat{d}=1|\textbf{x}) \big) \bigg] \\ 
    &= \mathop{\min}_{\theta_h} \mathop{\max}_{\theta_d} \
    \sum_{\textbf{x} \in X } \bigg[d \log F_d(F_h(\textbf{x})) + (1-d) \log \big( 1-F_d(F_h(\textbf{x})) \big) \bigg],
\end{aligned}
\end{equation} 
where $X=(\cup_{j=1}^{k} {X}^{S_j})\cup X^T$ and $d$ is the true domain (i.e., source or target) that $\textbf{x}$ belongs to.
The maximization objective in Equation~\ref{eq:minmax} promotes a more accurate domain classifier, while the minimization objective strengthens the domain invariance of the transformed feature vector $\textbf{h}$ by making hard even for the most accurate domain classifier to discern its domain identity.
Following \cite{ganin2016domain}, the two adversarial objectives in Equation~\ref{eq:minmax} can be unified using a gradient reversal layer $R(\textbf{z})$ defined as:
\begin{equation}
\label{eq:reverse layer}
    R(\textbf{z}) = \textbf{z}; \frac{dR(\textbf{z})}{d\textbf{z}} = -\textbf{I},
\end{equation} 
where $z$ denotes an input vector to the gradient reversal layer, and $\textbf{I}$ is an identity matrix. According to Equation~\ref{eq:reverse layer}, the gradient reversal layer makes no change to its input but flips the sign of the gradient passed through it. 
Equipped with the gradient reversal layer, solving the min-max problem in Equation~\ref{eq:minmax} is equivalent to minimize the following loss \citep{ganin2016domain}:
\begin{equation} 
\label{eq:da_up}
\begin{aligned}    
    \mathcal{L}_d(\theta_h, \theta_d) = - 
    \sum_{\textbf{x} \in X } \bigg[d \log F_d(R(F_h(\textbf{x}))) + (1-d) \log \big( 1-F_d(R(F_h(\textbf{x}))) \big) \bigg],
\end{aligned}
\end{equation} 
where we replace $F_h(\textbf{x})$ in Equation~\ref{eq:minmax} with $R(F_h(\textbf{x}))$.
To minimize the loss defined in Equation~\ref{eq:da_up}, parameters $\theta_d$ of the domain classifier $F_d$ are updated with gradient descent, leading to a more accurate domain classifier. In contrast, because of the gradient reversal layer, parameters $\theta_h$ of the feature transformation layer $F_h$ are adjusted with gradient upscent, resulting in more domain-invariant $\textbf{h}$ \citep{ganin2016domain}.

\subsection{Concept Alignment Module } \label{sec:4.4}

The key novelty of our method is the proposed concept alignment module, which mitigates concept shift between each source domain and the target domain. 
By Proposition~\ref{theorem:multi}, to mitigate concept shift, we need to learn a function that measures the similarity between two pieces of information.
To this end, we propose a similarity function submodule. 
Specifically, for two pieces of information respectively characterized by feature vectors $\textbf{x}_i$ and $\textbf{x}_j$, we define the similarity between them as
\begin{equation} \label{eq:sim}
F_s(\textbf{x}_i,\textbf{x}_j) = \frac{\text{MLP}_4(\textbf{x}_i;\theta_s) \cdot \text{MLP}_4(\textbf{x}_j;\theta_s)}{||\text{MLP}_4(\textbf{x}_i;\theta_s)|| \hspace{0.5mm} ||\text{MLP}_4(\textbf{x}_j;\theta_s)||}, 
\end{equation}
where $F_s$ denotes the similarity function, $\text{MLP}_4$ is an MLP with learnable parameters $\theta_s$ that transforms a feature vector (e.g., $\textbf{x}_i$) into a different representation space, notation $\cdot$ denotes the dot product of two vectors, and $||\hspace{0.1cm}||$ represents the $L^2$ norm of a vector. 
The objective of $\text{MLP}_4$ is to transform feature vectors into a representation space where information instances with the same label exhibit greater similarity than instances with different labels. This transformation facilitates concept alignment because differently labeled information instances (i.e., true information instances vs. false information instances) are better separated in the new representation space compared to the original one.

To learn the parameters $\theta_s$ of $\text{MLP}_4$, we employ a contrastive learning technique that maximizes the similarities between a focal information instance and its positive peers while minimizing the similarities with its negative peers. Peers can be defined in two ways \citep{le-khac_contrastive_2020}: the supervised way, which uses label information to identify positive and negative peers for a focal instance \citep{khosla_supervised_2020}, and the unsupervised way, which relies on heuristics to define peer relationships \citep{chen2020simple}. 
Given that information instances in source domains have labels, we opt for the supervised way.
Specifically, for each piece of source domain information characterized by feature vector $\textbf{x}$, we randomly select another piece of source domain information, bearing the same label, as its positive peer \citep{gunel2021supervised}. We denote the feature vector of the positive peer as $\textbf{x}^+$. In addition, we randomly select $m$ other pieces of source domain information, each having the opposite label of the focal information instance characterized by $\textbf{x}$, to serve as its negative peers \citep{chen2022dual}. We denote the feature vectors of these negative peers as $\textbf{x}_l^-$, $l=1,2,\dots,m$. The objective is to maximize the similarity $F_s(\textbf{x},\textbf{x}^+)$ between $\textbf{x}$ and $\textbf{x}^+$, while minimizing the similarity $F_s(\textbf{x},\textbf{x}_l^-)$ between $\textbf{x}$ and $\textbf{x}_l^-$, $l=1,2,\dots,m$. 
Accordingly, the parameters $\theta_s$ are learned by minimizing the InfoNCE loss commonly used by contrastive learning methods \citep{oord2018representation, chen2020simple}:  
\begin{equation} 
\label{eq:infonce}
    \mathcal{L}_s(\theta_s) = - \sum_{\textbf{x} \in X^S } \log \frac{\exp (F_s(\textbf{x},\textbf{x}^+) / \tau)}   {\exp (F_s(\textbf{x},\textbf{x}^+) / \tau) + \Sigma_{l=1}^m \exp(F_s(\textbf{x},\textbf{x}_l^-) / \tau)} ,
\end{equation}  
where $X^S=\cup_{j=1}^{k} {X}^{S_j}$, $F_s$ is given in Equation \ref{eq:sim}, and $0<\tau<1$ is a hyperparameter.\footnotemark\footnotetext{
Hyperparameter $\tau$ adjusts the emphasis on the two objectives \citep{frosst2019analyzing}. Lower $\tau$ enables the learning of the similarity function to place more emphasis on maximizing $F_s(\textbf{x},\textbf{x}^+)$ while higher $\tau$ shifts the learning to emphasize more on minimizing $F_s(\textbf{x},\textbf{x}_l^-)$.} By minimizing $\mathcal{L}_s$ w.r.t. $\theta_s$, this submodule learns $F_s$ that measures the similarity between two pieces of information in their transformed representation space such that identically labeled information instances are close to each other and distinctly labeled information instances are distant from each other.

Applying the similarity function $F_s$, for each piece of source domain information characterized by feature vector $\textbf{x}_i^{S_j}$, the concept shift mitigation submodule identifies its most similar target domain information characterized by feature vector $c(\textbf{x}_i^{S_j})$, $j=1,2,\dots,k$ and $i=1,2,\dots,N_{S_j}$. It then mitigates concept shift between source domains and the target domain by minimizing $\sum_{j=1}^k \sum_{i=1}^{N_{S_j}} |f_{S_j}(\textbf{x}_i^{S_j}) - f_T(c(\textbf{x}_i^{S_j}))|$ in Proposition~\ref{theorem:multi}. To this end, $f_{S_j}(\textbf{x}_i^{S_j})$ and $f_T(c(\textbf{x}_i^{S_j}))$ are estimated using the label classifier described in Section \ref{sec:lc_dc}. Specifically, we have 
\begin{equation*} 
\hat{f}_{S_j}(\textbf{x}_i^{S_j})=p(\hat{y}=1|\textbf{x}_i^{S_j})=F_y(F_h(\textbf{x}_i^{S_j})),
\end{equation*}  
where $\hat{f}_{S_j}$ is the estimation of $f_{S_j}$, and $F_h$ and $F_y$ are defined by Equations \ref{eq:fh} and \ref{eq:cls_layer}, respectively. Similarly, $f_T(c(\textbf{x}_i^{S_j}))$ is estimated by
\begin{equation*} 
\hat{f}_T(c(\textbf{x}_i^{S_j}))=p(\hat{y}=1|c(\textbf{x}_i^{S_j}))=F_y(F_h(c(\textbf{x}_i^{S_j}))).\footnotemark\footnotetext{The label classifier is trained to minimize the error of classifying source domain information. Thus, it is reasonable to estimate $f_{S_j}$ using the label classifier.
Since target domain information is unlabeled, we proxy $f_T$ using the label classifier, a strategy of which the effectiveness has been validated by previous studies \citep{
long2013transfer, 
long_conditional_2018, tachet_des_combes_domain_2020}.}
\end{equation*}
With the estimations of $f_{S_j}$ and $f_T$, the concept shift mitigation submodule reduces  concept shift by minimizing $\sum_{j=1}^k \sum_{i=1}^{N_{S_j}} |F_y(F_h(\textbf{x}_i^{S_j})) - F_y(F_h(c(\textbf{x}_i^{S_j})))|$.
Accordingly, this submodule is trained by minimizing the following loss:
\begin{equation} \label{eq:concept}
    \mathcal{L}_c(\theta_h, \theta_y) = \sum_{\textbf{x} \in X^S } 
    \Big[ 
    \big( F_y(F_h(\textbf{x})) - F_y(F_h(c(\textbf{x}))) \big)^2 - 
    \big( F_y(F_h(\textbf{x})) - F_y(F_h(d(\textbf{x}))) \big)^2 \Big],
\end{equation}
where $X^S=\cup_{j=1}^{k} {X}^{S_j}$. For each piece of source domain information with feature vector $\textbf{x} \in X^S$, $c(\textbf{x})$ and $d(\textbf{x})$
respectively denote the feature vectors characterizing its most similar and most dissimilar target domain information, both of which are identified using the similarity function $F_s$. Minimizing the first term in $\mathcal{L}_c$ essentially minimizes $\sum_{j=1}^k \sum_{i=1}^{N_{S_j}} |F_y(F_h(\textbf{x}_i^{S_j})) - F_y(F_h(c(\textbf{x}_i^{S_j})))|$. Consequently, it enforces the desired property that similar source and target domain information have similar label predictions \citep{liu2022deep}, thus reducing concept shift between source domains and the target domain. Minimizing the second term $-\big( F_y(F_h(\textbf{x})) - F_y(F_h(d(\textbf{x}))) \big)^2$ maximizes the label prediction difference between $F_y(F_h(\textbf{x}))$ and $F_y(F_h(d(\textbf{x})))$. As a result, label predictions diverge for dissimilar source and target domain information, which further mitigates concept shift.

\subsection{The DACA Method}\label{sec:4.5}
 

The DACA method is trained to minimize the combined losses of its three modules:  
\begin{equation} \label{eq:overall_obj}
\begin{aligned}
     \mathcal{L}_{\text{DACA}}(\theta_h, \theta_y, \theta_d, \theta_s) =
    \mathcal{L}_y(\theta_h, \theta_y) + \mathcal{L}_d(\theta_h, \theta_d) + \mathcal{L}_s(\theta_s) + \mathcal{L}_c(\theta_h, \theta_y)
\end{aligned}
\end{equation}
where $\mathcal{L}_y$, $\mathcal{L}_d$, $\mathcal{L}_s$, and $\mathcal{L}_c$ are defined by Equations \ref{eq:cls}, \ref{eq:da_up}, \ref{eq:infonce}, and \ref{eq:concept}, respectively. 
The losses $\mathcal{L}_y$ and $\mathcal{L}_d$ are associated with the classification module and the covariate alignment module, respectively. The losses $\mathcal{L}_s$ and $\mathcal{L}_c$ correspond to the similarity function and concept shift mitigation submodules of the concept alignment module, respectively.
There are two practical considerations when training DACA. First, the DACA method is trained in a two-stage manner. 
In the first stage (or the warmup stage), the method is trained by minimizing $\mathcal{L}_{\text{DACA}}-\mathcal{L}_c$. The rationale is that, to accurately estimate the concept alignment loss $\mathcal{L}_c$, DACA needs to learn a reliable similarity function first. 
In the second stage, DACA is trained by minimizing $\mathcal{L}_{\text{DACA}}$ as defined by Equation \ref{eq:overall_obj}. Second, to speed up the training of the DACA method, it is beneficial to mitigate concept shift between each pair of source domains. This is accomplished by minimizing a loss function for each pair of source domains. Concretely, for a pair of source domains $S_j$ and $S_n$, the loss function has the same form as Equation \ref{eq:concept} but with $S_j$ treated as the source domain in the equation and $S_n$ regarded as the target domain.
Once trained, DACA predicts the probability that a piece of target domain information characterized by feature vector $\textbf{x}_T$ is false as $F_y(F_h(\textbf{x}_T))$, where $F_h$ and $F_y$ are defined by Equations \ref{eq:fh} and \ref{eq:cls_layer}, respectively.

\section{Empirical Evaluation } \label{sec:empirical}

\subsection{Data} \label{sec:data}

\begin{figure*}[htbp]
\centering
\caption{Examples of True and Fake News in Evaluation Datasets }
\includegraphics[width=1\textwidth]{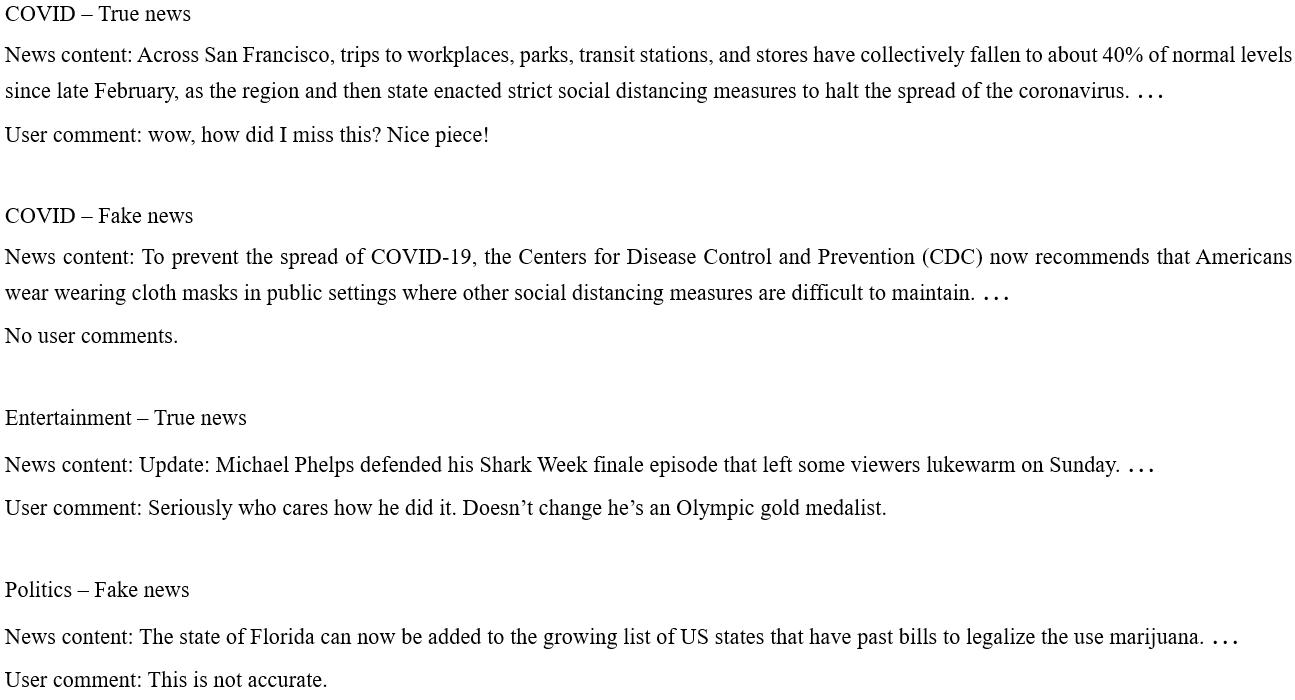}
\label{fig:example}
\end{figure*}

We evaluated the performance of our proposed DACA method using the publicly available datasets of English news, which have been widely employed to assess the performance of misinformation detection methods \citep{nan2021mdfend,mosallanezhad2022domain,zhu2022memory}. 
One is the MM-COVID dataset, which contains 4,750 pieces of true news (i.e., true information) and 1,317 pieces of fake news (i.e., misinformation) on COVID-19 as well as user comments on these news \citep{li2020mm}. Specifically, 8\% of COVID news are accompanied by user comments.
In our evaluation, we treated the COVID domain as the infodemic or target domain. 
For source domains, we utilized the FakeNewsNet dataset, which consists of true and fake news alongside their associated comments from the domains of entertainment and politics \citep{shu2020fakenewsnet}.\footnotemark\footnotetext{Both datasets were constructed by extracting true and fake news articles from reputable fact-checking websites, such as PolitiFact (\url{https://www.politifact.com}), where domain experts and professional journalists provide verified assessments on the veracity of news articles \citep{shu2020fakenewsnet, li2020mm}
Social media posts on X (formerly Twitter) containing links to these fact-checked news articles were collected and treated as user comments \citep{shu2020fakenewsnet, li2020mm}. While most users comment only once on an article, multiple comments on an article from the same user are occasionally observed.} Specifically, there are 16,804 pieces of true news and 5,067 pieces of fake news from the entertainment domain, and 1,583 instances of true news and 1,287 instances of fake news from the politics domain. User comments accompany 27\% of entertainment news and 59\% of politics news. The average comment length is 8.2 words in the COVID domain, 7.7 words in the entertainment domain, and 8.2 words in the politics domain.  
Table~\ref{tab:datastat} reports the summary statistics of the datasets used in our evaluation. 
Examples of true and fake news from the source and target domains are given in Figure~\ref{fig:example}.

\begin{table*}[htbp]
\centering
\caption{Summary Statistics of Evaluation Datasets. }
\def\arraystretch{1.3}\begin{tabular}{lccc}
\toprule 
Domain & Number of true news & Number of fake news \\
\midrule
COVID (Infodemic / Target)& 4,750 & 1,317  \\
Entertainment (Source) & 16,804 & 5,067  \\ 
Politics (Source) & 1,583 & 1,287  \\
\bottomrule 
\end{tabular}%
\label{tab:datastat}
\end{table*}%

\subsection{Evaluation Procedure and Benchmark Methods} \label{sec:setup}

During the early stage of an infodemic, there is little understanding of the disease that causes the infodemic. Consequently, even experts encounter difficulty distinguishing true news from fake news in the infodemic domain, leaving all news in this domain unlabeled. 
Hence, this scenario entails each method (ours or a benchmark) utilizing labeled news in the entertainment and politics domains to predict the label for each piece of news in the COVID domain.
Accordingly, the inputs to each method encompass labeled news in the entertainment and politics domains, along with user comments on these news, as well as \textit{unlabeled} news in the COVID domain and their associated comments. 
As the infodemic progresses, experts
gradually acquire more knowledge, leading to a small portion of the infodemic information being labeled. Consequently, we also consider scenarios of partially labeled infodemic information. 
Concretely, we compared our method with the benchmarks by increasing the number of labeled news articles in the COVID domain from $50$ to $150$, in increments of $50$. For example, each method utilizes labeled news in the entertainment and politics domains, together with $50$ labeled news articles in the COVID domain, to predict the label for each unlabeled news article in the COVID domain.

In each evaluation scenario, we conducted 25 experiments for every method (ours or a benchmark) and measured its performance using the metrics precision, recall, F1-score, and F2-score.
To curb the spread of fake news and minimize their societal impact, it is crucial to identify as many instances of fake news as possible \citep{zhu2022memory}. 
To this end, recall is an important metric as it measures the effectiveness of a method in identifying fake news. In addition, F1-score evaluates a method's performance in both identifying fake news and avoiding predicting true news as fake. Similarly, F2-score quantifies the performance in both precision and recall, with more weight on recall.
Concretely, let P be the number of fake news, TP be the number of fake news that are predicted as fake, and FP be the number of true news that are predicted as fake. For this case, precision is defined as $\text{TP}/(\text{TP} + \text{FP})$, recall is defined as $\text{TP}/\text{P}$, F1-score is computed as $2\text{TP}/(\text{TP}+\text{FP}+\text{P})$, and F2-score, which weights recall higher than precision, is given by $5\text{TP}/(\text{TP} + \text{FP} + 4\text{P})$.

As reviewed in Section~\ref{sec:rw} and summarized in Table \ref{tab:rw}, among all existing misinformation detection methods, only cross-domain misinformation detection methods with unlabeled target domain information can solve the EDM problem investigated in this paper. Therefore, we benchmarked our method against state-of-the-art methods in this category. Specifically, one benchmark is the MMD method, which minimizes covariate shift between source and target domain information, measured using the maximum mean discrepancy (MMD) metric \citep{huang2021dafd}. Another benchmark is the contrastive adaptation network for misinformation detection (CANMD) method, which assesses covariate shift using a variant of the MMD metric \citep{yue2022contrastive}. In addition, we also compared our method against  the domain adversarial neural network (DANN) method; it minimizes covariate shift by learning domain-invariant features from source and target domain information \citep{li2021multi}.
Moreover, general-purpose domain adaptation methods that were not originally designed for misinformation detection can also be adapted to solve our problem. As analyzed in Section~\ref{sec:rw:cross}, there are two categories of domain adaptation methods. One category explicitly measures and minimizes covariate shift while the other category implicitly minimizes covariate shift by learning domain-invariant features. Accordingly, we benchmarked against representative methods in each category. For the category that explicitly measures and minimizes covariate shift, we considered the multi-kernel MMD (MK-MMD) method \citep{long2015learning}, the joint adaptation network (JAN) method \citep{long2017deep}, and the discriminator-free adversarial learning network (DALN) method \citep{chen2022reusing} as our baselines. MK-MMD and JAN \citep{long2015learning,long2017deep} are commonly used domain adaption methods whereas DALN \citep{chen2022reusing} is a state-of-the-art method in this category. Representative methods in the other category include the maximum classifier discrepancy (MCD) method \citep{saito2018maximum} and the smooth domain adversarial training (SDAT) method \citep{rangwani2022closer}. 
MCD is widely used for domain adaptation and SDAT is a state-of-the-art implicit domain adaptation method. Table~\ref{tab:baseline} lists all methods compared in our evaluation.
One category of existing methods -- cross-domain misinformation detection method with labeled target domain information, which is not applicable to the scenario of completely unlabeled target domain information, becomes applicable to the scenarios of partially labeled target domain information. Therefore, we included REAL-FND \citep{mosallanezhad2022domain}, a representative method in this category, as an additional benchmark in these scenarios.

\begin{table*}[htbp]
\centering
\caption{Methods Compared in Our Evaluation. }
\def\arraystretch{1.3}
\begin{tabular}{ p{2cm} p{12cm} } 
\toprule 
Method & Notes  \\ 
\midrule
DACA & Our method \\ 
MMD & Cross-domain misinformation detection method based on the MMD metric \citep{huang2021dafd} \\ 
CANMD & Cross-domain misinformation detection method based on a variant of the MMD metric \citep{yue2022contrastive} \\
DANN & Cross-domain misinformation detection method based on the learning of domain-invariant features \citep{li2021multi} \\ 

MK-MMD & Domain adaptation method based on the multi-kernel MMD metric \citep{long2015learning} \\ 
JAN & Domain adaptation method based on the joint MMD metric \citep{long2017deep} \\ 
DALN & Domain adaptation method using the Nuclear-norm Wasserstein Discrepancy \citep{chen2022reusing} \\ 

MCD & Domain adaptation method utilizing a mini-max mechanism to align feature distributions between source and target domains \citep{saito2018maximum} \\ 
SDAT & Domain adaptation method employing a smoothing mechanism to learn domain-invariant features \citep{rangwani2022closer} \\ 
\bottomrule 
\end{tabular}%
\label{tab:baseline}
\end{table*}%

In each evaluation scenario, all compared methods took identical inputs and employed the same content embedding method  described in Section \ref{sec:4.1.2} to represent the textual contents of these inputs. 
We trained our DACA method using the Adam optimizer with a learning rate of $0.0001$ \citep{DBLP:journals/corr/KingmaB14}. The dimensions of the vectors $\mathbf{x}$ and $\mathbf{h}$ were set to 320.
The hyperparamters in Equation \ref{eq:infonce} of our method were set as follows: the number of instances of source domain information $m$ was set to 3, and the hyperparameter $\tau$ was set to 0.5.
Implementation details of the benchmark methods are given in Appendix~\ref{appendix:benchmark}.



\subsection{Evaluation Results} 
\label{sec:result}

\begin{table*}[htbp]
\centering
\caption{Performance Comparison between Our Method and Benchmark Methods (Zero Labeled News Articles in the COVID Domain).}
\def\arraystretch{1.3}
\begin{tabular} { p{2cm} p{2.5cm}<{\centering} p{2.5cm}<{\centering} p{2.5cm}<{\centering} p{2.5cm}<{\centering}} 
\toprule 
Method & Precision & Recall & F1-score & F2-score\\
\midrule
\multirow{2}{*}{DACA} & \textbf{0.716} & \textbf{0.745} & \textbf{0.719} & \textbf{0.739} \\ \addlinespace[-1ex]
& (0.030) & (0.049) & (0.034) & (0.043) \\ 
\multirow{2}{*}{MMD}  & 0.676$^{**}$ & 0.704$^{**}$ & 0.678$^{**}$ & 0.701$^{**}$ \\ \addlinespace[-1ex]
& (0.044) & (0.044) & (0.049) & (0.029) \\
\multirow{2}{*}{CANMD} & 0.655$^{**}$ & 0.702$^{**}$ & 0.661$^{**}$ & 0.692$^{**}$ \\ \addlinespace[-1ex]
& (0.028) & (0.026) & (0.030) & (0.027) \\
\multirow{2}{*}{DANN} & 0.639$^{**}$ & 0.658$^{**}$ & 0.650$^{**}$ & 0.656$^{**}$ \\ \addlinespace[-1ex]
& (0.035) & (0.043) & (0.030) & (0.040) \\
\multirow{2}{*}{MK-MMD} & 0.682$^{*}$ & 0.715$^{*}$ & 0.687$^{**}$ & 0.708$^{*}$ \\ \addlinespace[-1ex]
& (0.040) & (0.038) & (0.041) & (0.038) \\
\multirow{2}{*}{JAN} & 0.673$^{**}$ & 0.710$^{**}$ & 0.686$^{**}$ & 0.702$^{**}$ \\ \addlinespace[-1ex]
& (0.069) & (0.080) & (0.084) & (0.029) \\
\multirow{2}{*}{DALN} & 0.714 & 0.669$^{**}$ & 0.660$^{**}$ & 0.676$^{**}$ \\ \addlinespace[-1ex]
& (0.063) & (0.039) & (0.052) & (0.031) \\

\multirow{2}{*}{MCD} & 0.629$^{**}$ & 0.671$^{**}$ & 0.608$^{**}$ & 0.663$^{**}$ \\ \addlinespace[-1ex]
& (0.020) & (0.020) & (0.049) & (0.020) \\ 
\multirow{2}{*}{SDAT} & 0.675$^{**}$ & 0.712$^{**}$ & 0.682$^{**}$ & 0.705$^{**}$ \\ \addlinespace[-1ex]
& (0.015) & (0.002) & (0.015) & (0.002) \\
\bottomrule 
\end{tabular}%
\begin{tablenotes}
\centering\item[*] Note: Significance levels are denoted by * and ** for 0.05 and 0.01, respectively. 
\end{tablenotes}
\label{tab:improvement}
\end{table*}

Table~\ref{tab:improvement} presents the average precision, recall, F1-score, and F2-score for each method in the evaluation scenario of  completely unlabeled COVID domain, along with standard deviations (in parentheses), across 25 experiments.\footnotemark\footnotetext{For each method (ours or benchmark), we conducted 25 runs, each with a different random seed. To ensure a fair comparison, all methods used the same seed in each run.}
The t-test results show that our DACA method significantly outperforms each benchmark method in precision, recall, F1-score, and F2-score. In particular, our method achieves an average recall of 0.745.
Such performance is attained without any labeled COVID news in the training data, highlighting the efficacy of our method in transferring a model learned from labeled news in the entertainment and politics domains to predict the labels of news in the COVID domain. 
Moreover, our method respectively outperforms three state-of-the-art misinformation detection methods---MMD, CANMD, and DANN---by 5.92\%, 9.31\%, and 12.06\% in precision, by 5.82\%, 6.13\%, and 13.22\% in recall, by 6.05\%, 8.77\%, and 10.62\% in F1-score, and by 5.42\%, 6.79\%, and 12.65\% in F2-score. 
Additionally, the performance advantages of our method over representative domain adaptation methods---MK-MMD, JAN, DALN, MCD, and SDAT---range from 0.28\% to 13.83\% in precision, from 4.20\% to 11.36\% in recall, from 4.66\% to 18.26\% in F1-score, and from 4.38\% to 12.65\% in F2-score. We also compared the performance of our method with the benchmarks in terms of specificity. Our method significantly outperforms all benchmarks, except DALN, which achieves slightly higher specificity.
Since the key methodological difference between our method and the benchmark methods lies in the mitigation of concept shift by its concept alignment module, the performance improvements achieved by our method can be largely attributed to this module. Given the huge volume of information generated at the early stage of an infodemic, such performance advantages achieved by our method could result in substantially more instances of misinformation being identified by our method, in comparison to the benchmarks. As a result, a greater volume of misinformation could be prevented from dissemination, thereby significantly benefiting public health and society at large \citep{buchanan2020people, van2022misinformation}.

Tables~\ref{tab:body:partlabel50},~\ref{tab:body:partlabel100}, and~\ref{tab:body:partlabel150} present the average precision, recall, F1-score, and F2-score (with standard deviations in parentheses) for each method across 25 experiments, when the target COVID domain contains 50, 100, and 150 labeled news articles, respectively. 
As reported, our DACA method significantly outperforms each benchmark method in recall, F1-score, and F2-score, across varying numbers of labeled news articles in the target domain. It also surpasses each benchmark in precision, though the improvements in precision are not statistically significant in some cases. In addition, as the number of labeled news articles in the target domain increases, misinformation detection becomes easier, leading to improved performance across all methods. For example, our method achieves a recall of 0.745, 0.871, 0.911, and 0.928, when the target domain contains 0, 50, 100, and 150 labeled news articles, respectively.
Taken together, the evaluation results demonstrate the consistent effectiveness of our method in detecting misinformation as an infodemic progresses -- from its very early stage, when all infodemic information is completely unlabeled, to later stages, when partial information is labeled.\footnotemark\footnotetext{
To examine the boundary conditions under which the performance of our method and the benchmarks converges, we compared our method with the best-performing benchmark method (MK-MMD) under three scenarios of 10\%, 25\%, and 50\% labeled news articles in the COVID domain. The comparison results show that the recalls of the two methods converge when 50\% of the COVID news articles are labeled, while their F1-scores and F2-scores become similar when 25\% of the COVID news articles are labeled. However, it is unlikely that 25\% or more COVID news articles would be labeled at the early stage of an infodemic.}


\begin{table*}[htbp]
\centering
\caption{Performance Comparison between Our Method and Benchmark Methods (50 Labeled News Articles in the COVID Domain).}
\label{tab:body:partlabel50}
\def\arraystretch{1.1}
\begin{tabular}{ p{2cm} p{2.5cm}<{\centering} p{2.5cm}<{\centering} p{2.5cm}<{\centering} p{2.5cm}<{\centering} } 
\toprule 
Method & Precision & Recall & F1-score & F2-score \\
\midrule
\multirow{2}{*}{DACA} & \textbf{0.819} & \textbf{0.871} & \textbf{0.838} & \textbf{0.860} \\ \addlinespace[-1ex]
& (0.015) & (0.015) & (0.017) & (0.013) \\ 
\multirow{2}{*}{MMD} & 0.817 & 0.832$^{**}$ & 0.823$^{*}$ & 0.829$^{**}$ \\ \addlinespace[-1ex]
& (0.018) & (0.016) & (0.014) & (0.014) \\
\multirow{2}{*}{CANMD}  & 0.794$^{**}$ & 0.820$^{**}$ &  0.803$^{**}$ & 0.815$^{**}$  \\ \addlinespace[-1ex]
& (0.032) & (0.023) & (0.034) & (0.026) \\
\multirow{2}{*}{DANN}  & 0.812 & 0.803$^{**}$ & 0.803$^{**}$ & 0.804$^{**}$ \\ \addlinespace[-1ex]
& (0.030) & (0.028) & (0.019) & (0.021) \\

\multirow{2}{*}{MK-MMD}  & 0.785$^{**}$ & 0.811$^{**}$ &  0.795$^{**}$ & 0.805$^{**}$ \\ \addlinespace[-1ex]
& (0.027) & (0.019) & (0.023) & (0.020)\\

\multirow{2}{*}{JAN}  & 0.737$^{**}$ & 0.767$^{**}$ & 0.744$^{**}$ & 0.760$^{**}$ \\ \addlinespace[-1ex]
& (0.037) & (0.019) & (0.032) & (0.021)\\

\multirow{2}{*}{DALN}  & 0.770$^{**}$ & 0.831$^{*}$ & 0.799$^{*}$ & 0.817$^{**}$ \\ \addlinespace[-1ex]
& (0.044) & (0.031) & (0.032) & (0.035) \\

\multirow{2}{*}{MCD}  & 0.769$^{**}$ & 0.787$^{**}$ & 0.778$^{**}$ & 0.783$^{**}$ \\ \addlinespace[-1ex]
& (0.011) & (0.012) & (0.011) & (0.010)  \\ 
\multirow{2}{*}{SDAT}  & 0.742$^{**}$ & 0.766$^{**}$ & 0.754$^{**}$ & 0.761$^{**}$ \\ \addlinespace[-1ex]
& (0.008) & (0.001) & (0.009) & (0.009) \\

\multirow{2}{*}{REAL-FND} & 0.770$^{**}$ & 0.798$^{**}$ & 0.772$^{**}$ & 0.791$^{**}$ \\ \addlinespace[-1ex]
& (0.052) & (0.027) & (0.046) & (0.028)\\

\bottomrule 
\end{tabular}%
\begin{tablenotes}
\centering\item[*] Note: Significance levels are denoted by * and ** for 0.05 and 0.01, respectively. 
\end{tablenotes}
\end{table*}

\begin{table*}[htbp]
\centering
\caption{Performance Comparison between Our Method and Benchmark Methods (100 Labeled News Articles in the COVID Domain).}
\label{tab:body:partlabel100}
\def\arraystretch{1.1}
\begin{tabular}{ p{2cm} p{2.5cm}<{\centering} p{2.5cm}<{\centering} p{2.5cm}<{\centering} p{2.5cm}<{\centering} } 
\toprule 
Method & Precision & Recall & F1-score & F2-score \\
\midrule
\multirow{2}{*}{DACA} & \textbf{0.865} & \textbf{0.911} & \textbf{0.883} & \textbf{0.902}\\ \addlinespace[-1ex]
& (0.018) & (0.015) & (0.012) & (0.016) \\ 
\multirow{2}{*}{MMD} & 0.844$^{**}$ & 0.868$^{**}$ & 0.853$^{**}$ & 0.863$^{**}$ \\ \addlinespace[-1ex]
& (0.021) & (0.016) & (0.012) & (0.012) \\
\multirow{2}{*}{CANMD} & 0.843$^{**}$ & 0.847$^{**}$ & 0.844$^{**}$ & 0.846$^{**}$  \\ \addlinespace[-1ex]
& (0.019) & (0.018) &  (0.018) & (0.016)\\
\multirow{2}{*}{DANN}  & 0.859 & 0.859$^{**}$ & 0.857$^{**}$ & 0.858$^{**}$ \\ \addlinespace[-1ex]
& (0.024) & (0.017) & (0.015) & (0.015)\\

\multirow{2}{*}{MK-MMD} & 0.835$^{**}$ & 0.854$^{**}$ & 0.843$^{**}$ & 0.850$^{**}$ \\ \addlinespace[-1ex]
& (0.017) & (0.014) & (0.011) & (0.012) \\
\multirow{2}{*}{JAN}  & 0.814$^{**}$ & 0.839$^{**}$ & 0.823$^{**}$ & 0.834$^{**}$ \\ \addlinespace[-1ex]
& (0.029) & (0.019) & (0.022) & (0.018) \\
\multirow{2}{*}{DALN}  & 0.854 & 0.882$^{**}$ & 0.866$^{*}$ & 0.876$^{**}$ \\ \addlinespace[-1ex]
& (0.016) & (0.012) & (0.011) & (0.009) \\

\multirow{2}{*}{MCD}  & 0.781$^{**}$ & 0.878$^{**}$ & 0.801$^{**}$ & 0.857$^{**}$ \\ \addlinespace[-1ex]
& (0.024) & (0.018) & (0.032) & (0.019)  \\ 
\multirow{2}{*}{SDAT}  & 0.811$^{**}$ & 0.852$^{**}$ & 0.831$^{**}$ & 0.844$^{**}$ \\ \addlinespace[-1ex]
& (0.012) & (0.003) & (0.011) & (0.009) \\

\multirow{2}{*}{REAL-FND} & 0.840$^{**}$ & 0.835$^{**}$ & 0.836$^{**}$ & 0.836$^{**}$ \\ \addlinespace[-1ex]
& (0.028) & (0.023) & (0.019) & (0.020)\\
\bottomrule 
\end{tabular}%
\begin{tablenotes}
\centering\item[*] Note: Significance levels are denoted by * and ** for 0.05 and 0.01, respectively. 
\end{tablenotes}
\end{table*}

\begin{table*}[htbp]
\centering
\caption{Performance Comparison between Our Method and Benchmark Methods (150 Labeled News Articles in the COVID Domain).}
\label{tab:body:partlabel150}
\def\arraystretch{1.1}
\begin{tabular}{ p{2cm} p{2.5cm}<{\centering} p{2.5cm}<{\centering} p{2.5cm}<{\centering} p{2.5cm}<{\centering} } 
\toprule 
Method & Precision & Recall & F1-score & F2-score \\
\midrule
\multirow{2}{*}{DACA} & \textbf{0.888} & \textbf{0.928} & \textbf{0.904} & \textbf{0.919} \\ \addlinespace[-1ex]
& (0.014) & (0.013) & (0.011) & (0.014) \\ 
\multirow{2}{*}{MMD}  & 0.865$^{*}$ & 0.892$^{**}$ & 0.877$^{**}$ & 0.887$^{**}$ \\ \addlinespace[-1ex]
& (0.016) & (0.009) & (0.009) & (0.006) \\
\multirow{2}{*}{CANMD}  & 0.861$^{*}$ & 0.870$^{**}$ & 0.865$^{**}$ & 0.868$^{**}$  \\ \addlinespace[-1ex]
& (0.017) & (0.016) & (0.014) & (0.014) \\
\multirow{2}{*}{DANN}  & \textbf{0.888} & 0.883$^{**}$ & 0.884$^{*}$ & 0.884$^{**}$ \\ \addlinespace[-1ex]
& (0.014) & (0.017) & (0.011) & (0.014) \\

\multirow{2}{*}{MK-MMD}  & 0.853$^{**}$ & 0.878$^{**}$ & 0.863$^{**}$ & 0.873$^{**}$ \\ \addlinespace[-1ex]
& (0.021) & (0.010) & (0.012) & (0.008) \\
\multirow{2}{*}{JAN}  & 0.837$^{**}$ & 0.860$^{**}$ & 0.847$^{*}$ & 0.855$^{**}$ \\ \addlinespace[-1ex]
& (0.017) & (0.011) & (0.011) & (0.009) \\
\multirow{2}{*}{DALN}  & 0.872$^{**}$ & 0.888$^{*}$ & 0.870$^{*}$ & 0.899$^{**}$ \\ \addlinespace[-1ex]
& (0.014) & (0.006) & (0.010) & (0.006)  \\

\multirow{2}{*}{MCD}  & 0.785$^{**}$ & 0.880$^{**}$ & 0.830$^{**}$ & 0.859$^{**}$ \\ \addlinespace[-1ex]
& (0.018) & (0.007) & (0.015) & (0.008)   \\ 
\multirow{2}{*}{SDAT}  & 0.836$^{**}$ & 0.865$^{**}$ & 0.850$^{**}$ & 0.859$^{**}$ \\ \addlinespace[-1ex]
& (0.006) & (0.005) & (0.003) & (0.006)  \\

\multirow{2}{*}{REAL-FND} & 0.868$^{**}$ & 0.870$^{**}$ & 0.869$^{**}$ & 0.870$^{**}$ \\ \addlinespace[-1ex]
& (0.030) & (0.020) & (0.024) & (0.021)\\

\bottomrule 
\end{tabular}%
\begin{tablenotes}
\centering\item[*] Note: Significance levels are denoted by * and ** for 0.05 and 0.01, respectively. 
\end{tablenotes}
\end{table*}

\clearpage

As robustness checks, we conducted a series of supplementary experiments, which are detailed in Appendix \ref{appendix:robustness}. Specifically, we (i) evaluated our method on two additional infodemic datasets (Appendices \ref{appendix:chinese} and \ref{appendix:constraint}), (ii) altered the target domain from Infodemic to Finance (Appendix \ref{appendix:finance}), (iii) varied the forecasting objective from information veracity to diffusion virality (Appendix \ref{appendix:regression}), and (iv) compared the performance of our method with that of LLM-based approaches for misinformation detection (Appendix \ref{appendix:LLM}). 
Across all evaluation settings, the empirical results consistently confirm the superior performance of our method relative to the benchmark methods. Furthermore, Appendix \ref{appendix:sensitivity} presents sensitivity analyses: one examining the impact of incomplete source domain data on our method’s performance (Appendix \ref{appendix:incomplete_info}), another assessing the effect of source-target domain closeness on performance (Appendix \ref{appendix:heterogeneity_info}), and a third evaluating the performance of our method under a chronological setting (Appendix \ref{appendix:chro}).

\subsection{Performance Analysis} \label{sec: ablation_study}

Having demonstrated the superior performance of our method over the benchmarks, it is intriguing to delve deeper and analyze the factors contributing to its performance advantages. To this end, we conducted ablation studies to investigate the contribution of each novel design artifact of our DACA method to its performance. 
In particular, we focused on the key novelty of our method -- the concept alignment module.  As elaborated in Section \ref{sec:4.4}, it consists of
two submodules: the similarity function submodule and the concept shift mitigation submodule. The former represents instances of information in a transformed space, whereas the latter computes distances between these instances in the transformed space and mitigates concept shift.
To evaluate the overall contribution of the concept alignment module, we removed this module from the DACA method and designated the resulting method as W/o\_SF\_CSM(\underline{W}ith\underline{o}ut both the \underline{S}imilarity \underline{F}unction and the \underline{C}oncept \underline{S}hift \underline{M}itigation submodules). The performance difference between DACA and W/o\_SF\_CSM reveals the overall contribution of the concept alignment module to the performance of our method. 
To further investigate the functioning mechanism of the concept alignment module, we dropped its similarity function submodule but kept its concept shift mitigation submodule. We named the resulting method W/o\_SF(\underline{W}ith\underline{o}ut the \underline{S}imilarity \underline{F}unction submodule). The performance difference between DACA and W/o\_SF uncovers the role of the similarity function submodule in concept alignment.

\begin{table*}[htbp]
\centering
\caption{Ablation Study of DACA.}
\def\arraystretch{1.2}
\begin{tabular}{ p{2cm} p{2cm}<{\centering} p{2cm}<{\centering} p{2cm}<{\centering} p{2cm}<{\centering} } 
\toprule 
Method & Precision & Recall & F1-score & F2-score \\
\midrule
\multirow{2}{*}{DACA} & \textbf{0.716} & \textbf{0.745} & \textbf{0.719} & \textbf{0.739} \\ \addlinespace[-1ex]
& (0.030) & (0.049) & (0.034) & (0.043) \\ 
\multirow{2}{*}{W/o\_SF\_CSM} & 0.650$^{**}$ & 0.658$^{**}$ & 0.650$^{**}$ & 0.656$^{**}$ \\ \addlinespace[-1ex]
& (0.047) & (0.053) & (0.047) & (0.056) \\ 
\multirow{2}{*}{W/o\_SF} & 0.660$^{**}$ & 0.663$^{**}$ & 0.648$^{**}$ & 0.662$^{**}$ \\ \addlinespace[-1ex]
& (0.049) & (0.047) & (0.053) & (0.050) \\ 
\bottomrule 
\end{tabular}%
\begin{tablenotes}
\centering\item[*] Note: Significance levels are denoted by * and ** for 0.05 and 0.01, respectively. 
\end{tablenotes}
\label{tab:ablation1}
\end{table*}


Table~\ref{tab:ablation1} presents the average performance of DACA, W/o\_SF\_CSM, and W/o\_SF in the evaluation scenario of completely unlabeled COVID domain, along with their respective standard deviations (in parentheses), across 25 experiments. 
As reported, the performance of W/o\_SF\_CSM is significantly inferior to that of DACA, due to the removal of the concept alignment module. 
More specifically, W/o\_SF\_CSM trails DACA with precision decreased from 0.716 to 0.650, recall decreased from 0.745 to 0.658, F1-score decreased from 0.719 to 0.650, and F2-score decreased from 0.739 to 0.656.
which collectively demonstrate the contribution of the concept alignment module to the performance of DACA. Furthermore, W/o\_SF performs significantly worse than DACA but comparable to W/o\_SF\_CSM. It is noted that W/o\_SF has the concept shift mitigation submodule while omitting the similarity function submodule. The comparison results thus show that, to realize the benefits of the concept shift mitigation submodule, it must be coupled with the similarity function submodule. This empirical finding is in line with the theoretical analysis in Section \ref{sec:4.1}, which defines concept shift over source domain instances and their corresponding nearest target domain instances. 
Therefore, to mitigate concept shift, we need a suitable similarity function that measures similarities between instances in an appropriate representation space,
which is realized by the similarity function submodule. We also evaluate the relative contributions of the covariate alignment and concept alignment modules by removing each module individually from DACA. The results show that the concept alignment module contributes more to the performance of DACA than the covariate alignment module. 
Overall, the ablation study demonstrates the significant contribution of the concept alignment module to the performance of our method. Moreover, the ablation study shows that it is appropriate to design this module with two submodules: similarity function and concept shift mitigation.

\section{Conclusion}

\subsection{Summary and Contributions  }

To contain harmful effects of an infodemic on public health, it is crucial to detect misinformation at the early stage of the infodemic. 
An early stage infodemic is characterized by a large volume of unlabeled information spread across various media platforms. Consequently, conventional misinformation detection methods are not suitable for this misinformation detection task because they rely on labeled information in the infodemic domain to train their models. State-of-the-art misinformation detection methods learn their models with labeled information in other domains to detect misinformation in the infodemic domain, thereby applicable to the task. The efficacy of these methods depends on their ability to mitigate both covariate shift and concept shift between the infodemic domain and the domains from which they leverage labeled information. These methods focus on mitigating covariate shift but overlook concept shift, making them less effective for the task. In response, we propose a novel misinformation detection method that addresses both covariate shift and concept shift. Through extensive empirical evaluations with widely used datasets, we demonstrate the superior performance of our method over state-of-the-art misinformation detection methods as well as prevalent domain adaptation methods that can be tailored to solve the misinformation detection task.


Our study makes the following contributions to the extant literature. First, our study belongs to the area of computational design science research in the IS field \citep{rai2017editor, padmanabhan2022machine, fang2025computational}. This area of research develops computational algorithms and methods to solve business and societal problems and aims at making methodological contributions to the literature, e.g., \cite{abbasi2010detecting, li2017utility, zhao2022exploiting}. In particular, the methodological contribution of our study lies in its addressing of concept shift, in addition to covariate shift. More specifically, we theoretically show the importance of addressing concept shift and how to operationalize it. Built on the theoretical analysis, we develop a novel concept alignment module to mitigate concept shift, as described in Section \ref{sec:4.4}. Second, given its significant social and economic impact, misinformation detection and management has attracted attention from IS scholars, e.g., \cite{moravec2019fake, wei2022combining, hwang2025nudge}. Our study adds to this stream of IS research with a novel method that is effective in detecting misinformation at the early stage of an infodemic.


\subsection{Implications for Infodemic Management and Future Work}
\label{sec: implication}

Every epidemic is accompanied by an infodemic, a phenomenon known since the Middle Ages \citep{zarocostas2020fight}. The wide dissemination of misinformation during an infodemic misleads people to dismiss health guidance and pursue unscientific treatments, resulting in substantial harm to public health and significant social and economic consequences \citep{bursztyn2020misinformation,romer2020conspiracy}. Furthermore, the pervasive reach of the Internet and social media platforms accelerates the spread of misinformation and amplifies its harmful impacts on public health and society \citep{zarocostas2020fight}. 
Detecting misinformation at the early stage of an infodemic discourages people from believing and sharing it, thereby preventing it from going viral \citep{buchanan2020people}. Hence, early detection of misinformation is crucial for managing an infodemic and mitigating its adverse effects. Accordingly, a direct implication of our study is to provide an effective early misinformation detection method for infodemic management. Our method effectively overcomes obstacles to early detection of misinformation. First, vast amount of information spread during the early stage of an infodemic makes manual identification of misinformation impractical. In response, our deep learning-based method automatically learns to differentiate misinformation from true information. Second, there is no labeled information at the early stage of an infodemic, rendering traditional misinformation detection methods inapplicable. Accordingly, our method leverages labeled information in other domains to detect misinformation in the infodemic domain. 

Our study empowers infodemic management in several ways. Operators of media platforms can employ our method to effectively detect misinformation in the infodemic domain that spreads on their platforms. Subsequently, they can flag and debunk identified misinformation, which helps curb the diffusion of misinformation and alleviate its negative impacts  \citep{pennycook2020implied,wei2022combining}. 
Additionally, flagged misinformation is less likely to be clicked, thereby reducing revenues for those who monetize misinformation and diminishing their incentives to create more misinformation
\citep{pennycook2021psychology}. Moreover, platform operators can trace misinformation identified by our method back to its producers. They could then restrict those who frequently produce misinformation from publishing on their platforms. In this regard, platforms are encouraged to set up guidelines to regulate their content producers \citep{hartley2020fighting}. Platform operators can also trace misinformation detected by our method to its spreaders. Understandably, blocking individuals who disseminate a large amount of misinformation is important for combating misinformation on their platforms.  

Furthermore, our study sheds light on the value of cross-domain data sharing for infodemic management. At the early stage of an infodemic, even experts have difficulty in distinguishing between misinformation and true information. Moreover, recruiting experts to verify and label information is costly \citep{kim2018leveraging}. Cross-domain data sharing enables us to utilize labeled information in other domains to detect misinformation in the infodemic domain. Hence, it is a viable and cost-effective approach to misinformation detection and infodemic management. To implement this approach, we need to address a challenge. Information from different domains exhibits different marginal and conditional distributions, known as covariate shift and concept shift, respectively. Thus, the challenge is how to mitigate covariate shift and concept shift between the infodemic domain and source domains that provide labeled information. Our proposed DACA method is effective in tackling this challenge. To facilitate cross-domain data sharing for infodemic management, media platforms are recommended to establish data exchange mechanisms to ensure that participating platforms are properly incentivized and labeled information is securely shared.

Our study has limitations and can be extended in several directions. First, our method employs information content to detect misinformation. Future work could extend our method by incorporating information propagation patterns. Specifically, for each piece of information, a diffusion graph can be constructed in which nodes represent users and edges represent the spread the information from one user to another. Next, a Graph Neural Network (GNN) can be employed to learn graph-level representations of these diffusion graphs. Concretely, for each diffusion graph, a GNN computes node embeddings, which are then pooled (e.g., via mean pooling) to obtain the embedding of the graph.
Finally, the resulting graph embeddings are concatenated with their corresponding information content embeddings and fed into our method. 
Moreover, while our method utilizes user comments by concatenating them with news content, advanced approaches could be developed to more effectively capture the signals in user comments for misinformation detection.
Second, although our method is designed for completely unlabeled target domains, we have empirically demonstrated how its performance evolves as labels in a target domain gradually become available. 
Future research could investigate semi-supervised or active learning extensions of our method for partially labeled target domains. 
Third, it is interesting to incorporate our misinformation detection method into a fact-checking system. This direction involves aligning our domain adaptation framework with pipelines that retrieve and verify evidence in real time, potentially improving the interpretability and accountability of the misinformation detection process. Finally, our method can be
generalized to solve domain adaptation problems in other contexts. Accordingly, broader evaluations across diverse problem contexts would further assess its generalizability and robustness.

\section{Acknowledgments}

The authors thank the department editor, the associate editor, and three anonymous reviewers for their guidance and constructive comments, which have greatly improved the paper. X. Zhao and X. Fang are corresponding authors. X. Fang is partially supported by the Lerner College of Business and Economics Small Research Grant. X. Zhao is partially supported by the National Natural Science Foundation of China (Grant No. 72401172).

\bibliographystyle{informs2014} 
\bibliography{reference} 
\endthebibliography

\newpage 





\setcounter{figure}{0}
\setcounter{table}{0}
\setcounter{section}{0}
\setcounter{equation}{0}
\setcounter{page}{1}

\renewcommand{\thetable}{A\arabic{table}}
\renewcommand{\thefigure}{A\arabic{figure}}
\renewcommand{\theequation}{A\arabic{equation}}
\renewcommand{\thepage}{EC\arabic{page}}

\begin{appendices}

\section{Empirical Evidence on Data Distribution Shift}
\label{appendix:dist_shift}

We empirically demonstrate the presence of covariate shift and concept shift between source and target domains. 
Let data in the source domain $S$ follow the joint distribution $p_{S}(\textbf{x},y)$ for $\textbf{x}\in\mathcal{X}$ and $y\in\mathcal{Y}$, where $\mathcal{X}$ denotes the feature space and $\mathcal{Y}$ is the label space. Similarly, data in the target domain $T$ are drawn from the joint distribution $p_{T}(\textbf{x},y)$. These joint distributions can be factorized as $p_{S}(\textbf{x},y)=p_S(\textbf{x}) p_S(y|\textbf{x})$ and $p_{T}(\textbf{x},y)=p_T(\textbf{x}) p_T(y|\textbf{x})$, respectively. 
To instantiate the source and target domains, we employ two real-world datasets. For the source domain (entertainment), we employed the FakeNewsNet dataset \citepec{shu2020fakenewsnet_ec}, which contains 16,804 true entertainment news pieces and 5,067 fake entertainment news pieces. For the target domain (infodemic), we utilized the MM-COVID dataset \citepec{li2020mm_ec}, which consists of 4,750 true news pieces and 1,317 fake news pieces on COVID-19.

\begin{figure}[htbp]
\FIGURE{\includegraphics[width=0.8\textwidth]{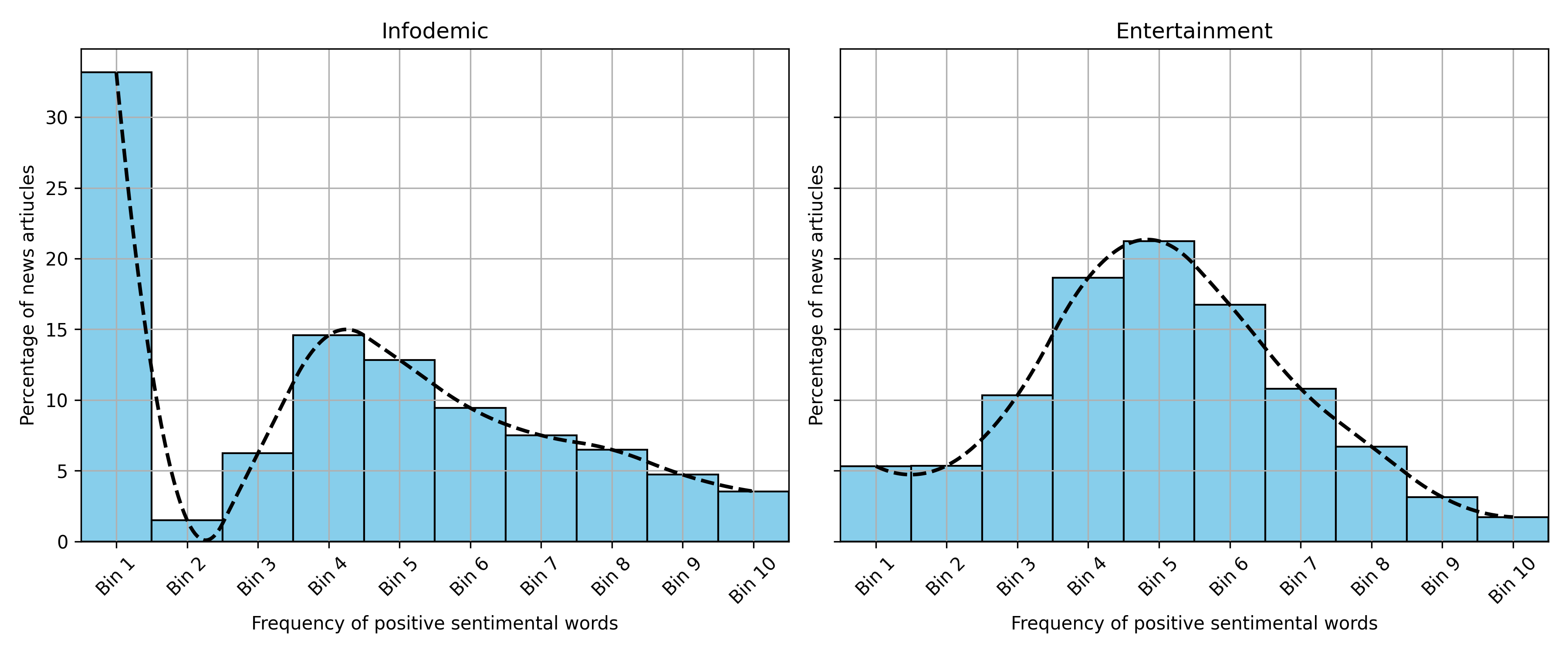}}
{Covariate Shift between Infodemic and Entertainment Domains
\label{fig:ec:covariate}}
{Better view in color.}
\end{figure}

According to \citeec{liu2022deep_ec}, covariate shift refers to the difference between the feature distribution in the source domain (i.e., $p_S(\textbf{x})$) and that in the target domain (i.e., $p_T(\textbf{x})$).
Previous studies have employed emotion features, such as the frequency of positive sentimental words and the frequency of exclamation marks in a news article, to distinguish between true and fake news \citepec[e.g.,][]{zhang2021mining_ec, zhu_memory-guided_2023_2}. 
Accordingly, we demonstrated covariate shift between the infodemic and entertainment domains by showing their difference in the distribution of the frequency of positive sentimental words.
\citeec{zhu_memory-guided_2023_2} have calculated that frequency for each news article in both domains and normalized the frequencies. We further discretized the range of normalized frequencies into 10 equal-range bins, with larger bin ID indicating higher frequency of positive sentimental words.
For each domain, we counted the number of new articles falling into each bin and plotted its proportion relative to the total number of news articles in that domain in Figure~\ref{fig:ec:covariate}. As shown, the distribution of the frequency of positive sentimental words in the infodemic domain is skewed toward bin 1 (i.e., the lowest frequency among the 10 bins), while the distribution in the entertainment domain exhibits a bell-shaped curve. For example, according to Figure~\ref{fig:ec:covariate},
news articles with the lowest frequency of positive sentimental words account for 33.18\% of all infodemic news articles, while the corresponding proportion in the entertainment domain is only 5.37\%.

\begin{figure}[htbp]
\FIGURE{\includegraphics[width=0.8\textwidth]{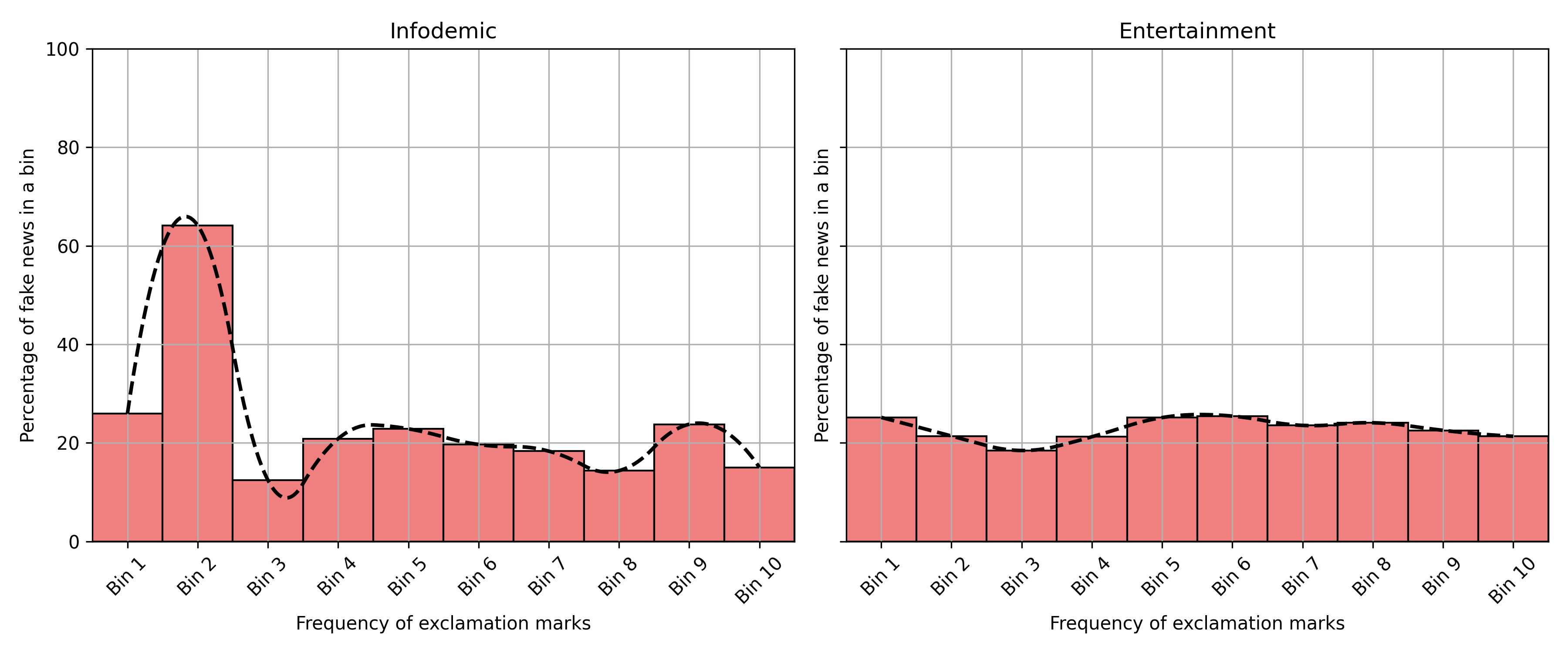}}
{Concept Shift between Infodemic and Entertainment Domains
\label{fig:ec:concept}}
{Better view in color.}
\end{figure}

Concept shift refers to the difference between $p_S(y|\textbf{x})$ in the source domain and $p_T(y|\textbf{x})$ in the target domain \citepec{liu2022deep_ec}. Since both $p_S(y|\textbf{x})$ and $p_T(y|\textbf{x})$ condition on features $\textbf{x}$, we employed the frequency of exclamation marks as a feature for fake news detection \citepec{zhang2021mining_ec, zhu_memory-guided_2023_2}.
\citeec{zhu_memory-guided_2023_2} have calculated this frequency for each news article in both infodemic (target) and entertainment (source) domains and normalized the frequencies. We then discretized the range of normalized frequencies into 10 equal-range bins, with a larger bin ID indicating a higher frequency of exclamation marks.
For each domain, we counted the number of \textit{fake} new articles as well as the number of all news articles falling into each bin and plotted the percentage of \textit{fake} new articles in each bin in Figure~\ref{fig:ec:concept}.
The figure illustrates concept shift between the two domains, as the label distribution conditioned on the emotion feature differs across domains. Specifically, the percentages of fake news surge in bin 2 for the infodemic domain, while they are stable across bins in the entertainment domain. The percentage of fake news in a bin generally indicates the likelihood that a news article is fake, given that the frequency of exclamation marks in the article falls into that bin.
For example, when conditioned on the feature value falling into bin 2, the likelihood of an infodemic news article being fake is about 64\%, while that for an entertainment news article being fake is around 21\%.

\section{Method Comparison in More Dimensions}
\label{appendix:compare}

In Table \ref{tab:ec:misinfo_summary}, we compare our method against existing misinformation detection methods along five dimensions: (1) Domain, which indicates whether a method leverages single domain, multi-domain, or cross-domain data; (2) Methodological Contribution, which  summarizes the core innovation of a method; (3) Types of Input Data, which lists the types of data utilized for model training; (4) Training Temporality, which describes the timing of model training in terms of whether it requires labeled samples from the target domain; and (5) Inference Temporality, which specifies when the model is capable of making predictions, i.e., whether it can generate predictions immediately upon content release or must wait for additional signals like diffusion information or user engagement data. These dimensions were chosen to highlight the practical constraints of model deployment at different stages of an infodemic, particularly in the early phase, when time-sensitive detection using completely or partially unlabeled target domain data is critical \citepec{li2021multi2, yue2022contrastive2}. The methods compared in the table are representative methods in each category of misinformation detection methods reviewed in Section \ref{sec:rw}. Table \ref{tab:ec:misinfo_summary} highlights our method’s advantages in real-world settings, particularly in terms of inference immediacy and minimal reliance on labeled target domain data.

\begin{center}
\small
\begin{longtable}{P{2.1cm} | P{1.2cm} | P{3.0cm} | P{2.8cm} |  P{2.8cm} | P{2.8cm}  }
\caption{Comparison of Misinformation Detection Methods}
\label{tab:ec:misinfo_summary} \\
\hline
 Method & Domain & Methodological Contribution & Types of Input Data  & Training Temporality & Inference Temporality  \\
\hline
\endfirsthead
\hline
Method & Domain & Methodological Contribution & Types of Input Data & Training Temporality & Inference Temporality  \\
\hline
\endhead

\citeec{wei_combining_2022_ec} & Single & Leverage crowd intelligence via Bayesian result aggregation & News content (text, presence of images, videos or external links), user engagement data (human responses and reports) & Need sufficient time to label enough target domain samples & Need additional time for observing user engagement data  \\
\hline
\citeec{sun_rumor_2022_ec} & Single & Counter noise and adversarial samples based on graph adversarial contrastive learning & News content (text), diffusion information (a tree connecting the original post with propagation events such as retweets and comments) & Need sufficient time to label enough target domain samples & Need additional time for collecting diffusion data \\
\hline
\citeec{zhang_reinforced_2024_ec} & Single & Select relevant external knowledge subgraph with reinforcement learning, and perform modality integration with attentive hierarchical pooling & Multimodal news content (text and images), external knowledge & Need sufficient time to label enough target domain samples & Immediate prediction at the point of content release  \\
\hline
\citeec{zhu_memory-guided_2023_2} & Multiple & Memory-guided multi-view and multi-domain integration based on a novel attention mechanism & News content (text, emotion features, style features)  & Need sufficient time to label enough target domain samples  & Immediate prediction at the point of content release \\
\hline
\citeec{lu_dammfnd_2025_ec} & Multiple & Employ domain disentanglement to extract domain invariant and specific representations from multimodal data & Multimodal news content (text and images)  & Need sufficient time to label enough target domain samples & Immediate prediction at the point of content release \\
\hline
\citeec{mosallanezhad2022domain_ec} & Cross & Introduce a reinforcement learning agent to generate domain invariant news representations & News content (text), user engagement data (comments, likes, or shares) & Need sufficient time to label enough target domain samples & Need additional time for observing user engagement data \\
\hline
\citeec{nan2022improving_ec} & Cross & Transfer knowledge at both domain and instance levels using meta-learning and language perplexity & News content (text)  & Need sufficient time to label enough target domain samples & Immediate prediction at the point of content release \\
\hline
\citeec{li2021multi2} & Cross & Domain adaptation with weak-label heuristics to utilize unlabeled target data, only addressing covariate shift & News content (text) & No need to wait for target domain labels & Immediate prediction at the point of content release \\
\hline
\citeec{yue2022contrastive2} & Cross & Contrastive domain adaptation with pseudo-labeling and label correction, only addressing covariate shift & News content (text) & No need to wait for target domain labels & Immediate prediction at the point of content release  \\
\hline
Our Method & Cross & A novel method that mitigates both covariate shift and concept shift & News content (text, emotion, writing styles), user comments (optional) & No need to wait for target domain labels & Immediate prediction at the point of content release \\
\hline
\end{longtable}
\end{center}

\section{$\mathcal{H}$-divergence } \label{appendix:definitions}

$\mathcal{H}$-divergence has been widely utilized to measure distances between distributions over a feature space.

\textbf{Definition A.1. ($\mathcal{H}$-divergence)} \citepec{ben2010theory2}. Given a  feature space $\mathcal{X}$, let $\mathcal{H}$ denote a hypothesis space defined on $\mathcal{X}$ and $\mathbb{D}_S$ and $\mathbb{D}_T$ be two probability distributions over $\mathcal{X}$. For a hypothesis $h \in \mathcal{H}$, let $I(h)$ be the set such that $\textbf{x}\in I(h)$ if and only if $h(\textbf{x}) > \tau$, where $\tau$ is the classification threshold and an instance with features $\textbf{x}$ is classified as 1 if $h(\textbf{x}) > \tau$ and 0 otherwise. The $\mathcal{H}$-divergence between $\mathbb{D}_S$ and $\mathbb{D}_T$ is defined as 
\begin{equation}
    d_{\mathcal{H}} (\mathbb{D}_S, \mathbb{D}_T) = 2 \sup_{h\in \mathcal{H}} |Pr_{\mathbb{D}_S}[I(h)]- Pr_{\mathbb{D}_T}[I(h)]|,
\end{equation}
where $Pr$ denotes probability.\footnotemark\footnotetext{\citeec{ben2010theory2} consider a hypothesis $h$ that produces a discrete output, i.e., 0 or 1, while we consider a hypothesis $h$ that predicts a continuous output between 0 and 1.} 

The $\mathcal{H}$-divergence between two distributions can be empirically estimated using two sample datasets drawn from these distributions, respectively \citepec{ben2010theory2}. Given a dataset of $N_S$ source domain instances $\mathcal{D}_S$ and a dataset of $N_T$ target domain instances $\mathcal{D}_T$, features of which are from distributions $\mathbb{D}_S$ and $\mathbb{D}_T$ respectively, the $\mathcal{H}$-divergence between $\mathbb{D}_S$ and $\mathbb{D}_T$ can be estimated using $\hat{d}_{\mathcal{H}}(\mathcal{D}_S, \mathcal{D}_T)$ \citepec{ben2010theory2}: 
\begin{equation} \label{eq:em_hdiver}
    \hat{d}_{\mathcal{H}}\left(\mathcal{D}_S, \mathcal{D}_T\right)=
    2 \left(1-\min _{h \in \mathcal{H}}\left[\frac{1}{N_S} \sum_{\mathbf{x}: h(\mathbf{x}) \leq \tau} I[\mathbf{x} \in \mathcal{D}_S] + \frac{1}{N_T} \sum_{\mathbf{x}: h(\mathbf{x}) > \tau} I\left[\mathbf{x} \in \mathcal{D}_T \right]\right]\right),
\end{equation}
where indicator function $I[\mathbf{x} \in \mathcal{D}_S ]$ is 1 if an instance with feature $\textbf{x}$ belongs to $\mathcal{D}_S $ and $\mathcal{H}$ is a symmetric hypothesis space (i.e., for each hypothesis $h\in\mathcal{H}$, its inverse $1-h$ is also in $\mathcal{H}$). 

\section{Proof of Theorem and Proposition } \label{appendix:proof1}


Let $\epsilon_S(g_1,g_2)= E_{\textbf{x}\sim \mathbb{D}_S} [|g_1(\textbf{x})-g_2(\textbf{x})| ]$ be the expected difference between $g_1$ and $g_2$ in the source domain, where $g_1$ and $g_2$ are two labeling functions. Similarly, $\epsilon_T(g_1,g_2)= E_{\textbf{x}\sim \mathbb{D}_T} [|g_1(\textbf{x})-g_2(\textbf{x})| ]$ denotes the expected difference between $g_1$ and $g_2$ in the target domain. To prove Theorem~\ref{theorem:single}, we first show an upper bound of $\epsilon_S\left(f_S,f_T\right)$:

\textbf{Lemma 1 }\textit{Let $f_S$ and $f_T$ respectively denote the true source and target domain labeling functions, $\mathbb{D}_S$ be the source domain feature distribution, and $\mathcal{D}_S$ be a dataset with $N_S$ samples drawn from $\mathbb{D}_S$, for any $\eta\in (0,1)$, 
the following inequality holds with probability at least $1-\eta $:} 
\begin{equation}
\begin{aligned} \label{eq:lemma1}
\epsilon_S\left(f_S,f_T\right)=E_{\textbf{x}\sim \mathbb{D}_S}[|f_S(\textbf{x})-f_T(\textbf{x})|] & \leq \frac{1}{N_S} \sum_{i=1}^{N_S} |f_S(\textbf{x}_i^S) - f_T(\textbf{x}_i^S) |  + \sqrt{-\frac{\ln(\frac{\eta}{2})}{2 N_S} }, \\ 
\end{aligned}
\end{equation}
\textit{where} $\textbf{x}_i^{S}$ \textit{is the $i$-th sample drawn i.i.d. from $\mathbb{D}_S$. }

\textit{Proof. This proof relies on McDiarmid's inequality \citepec{doob1940regularity2}. For notation convenience, define} $\textbf{z}=f_S(\textbf{x})-f_T(\textbf{x})$ \textit{where} $\textbf{x}\sim \mathbb{D}_S$. \textit{We then have} $\textbf{z}_i = f_S(\textbf{x}_i^{S})-f_T(\textbf{x}_i^{S}), i=1,2,\cdots,N_s$.  \textit{Further define} $g(\textbf{z}_1,\textbf{z}_2,\dots,\textbf{z}_{N_S}) = \frac{1}{N_S} \sum_{i=1}^{N_S} |\textbf{z}_i|$. \textit{Function $g$ satisfies the property that}
\begin{equation}
\begin{aligned}
|g(\textbf{z}_1,\textbf{z}_2,\dots,\textbf{z}_{N_S})-g(\textbf{z}_1,\dots,\textbf{z}_{k-1},\textbf{z}_k',\textbf{z}_{k+1}, \dots, \textbf{z}_{N_S})| \leq \frac{1}{N_S},
\end{aligned}
\end{equation}
\textit{which means that by changing any $k$-th coordinate, the function will change at most $\frac{1}{N_S}$. This is because} $|\textbf{z}_k|=|f_S(\textbf{x}_k^S)-f_T(\textbf{x}_k^S)|\in [0,1]$.  

\textit{Given that} $\textbf{z}_1,\textbf{z}_2,\dots,\textbf{z}_{N_S}$ \textit{are i.i.d. random samples, for any $\epsilon > 0$, by McDiarmid's inequality}
\begin{equation}
\begin{aligned}
p(|g(\textbf{z}_1,\textbf{z}_2,\dots,\textbf{z}_{N_S})-E [g(\textbf{z}_1,\textbf{z}_2,\dots,\textbf{z}_{N_S})] | \geq \epsilon)  & \leq 2 \exp (\frac{- 2\epsilon^2}{\sum_{k=1}^{N_S}\frac{1}{N_S^2}}) \\ 
& =  2 \exp (- 2 N_S \epsilon^2 ). 
\end{aligned}
\end{equation}
\textit{Equivalently,}
\begin{equation}
\begin{aligned}
p(|g(\textbf{z}_1,\textbf{z}_2,\dots,\textbf{z}_{N_S})-E [g(\textbf{z}_1,\textbf{z}_2,\dots,\textbf{z}_{N_S})] |  \leq \epsilon)  & \geq 1 - 2\exp (- 2 N_S \epsilon^2 ). 
\end{aligned}
\end{equation}
\textit{By replacing $\epsilon$ with $\sqrt{-\frac{\ln(\frac{\eta}{2})}{2 N_S} }$ where $\eta\in(0,1)$, we have}
\begin{equation}
\begin{aligned} \label{eq:a7}
p \Bigg( |g(\textbf{z}_1,\textbf{z}_2,\dots,\textbf{z}_{N_S})-E [g(\textbf{z}_1,\textbf{z}_2,\dots,\textbf{z}_{N_S})] |  \leq \sqrt{-\frac{\ln(\frac{\eta}{2})}{2 N_S} } \Bigg) & \geq  1-\eta ,
\end{aligned}
\end{equation}

\textit{Note that} $E [g(\textbf{z}_1,\textbf{z}_2,\dots,\textbf{z}_{N_S})] = E_{\textbf{x}\sim \mathbb{D}_S}[|f_S(\textbf{x})-f_T(\textbf{x})|]$ \textit{by construction, while} $g(\textbf{z}_1,\textbf{z}_2,\dots,\textbf{z}_{N_S}) = \frac{1}{N_S} \sum_{i=1}^{N_S} |f_S(\textbf{x}_i^S) - f_T(\textbf{x}_i^S) |$ \textit{by definition. Based on the inequality $a-b \le |b-a|$, 
Inequality \ref{eq:a7} means that Inequality \ref{eq:lemma1} must hold with probability at least $1-\eta$, which concludes our proof of Lemma 1.} 

Next, we show, in Lemma 2, that the population $\mathcal{H}$-divergence is bounded by the empirical $\mathcal{H}$-divergence. 

\textbf{Lemma 2 } \textit{Let $\mathcal{H}$ be a hypothesis space with VC dimension $d$. $\mathcal{D}_S$ and $\mathcal{D}_T$ are datasets of instances drawn from distributions $\mathbb{D}_S$ and $\mathbb{D}_T$, respectively. Let $d_\mathcal{H}(\mathbb{D}_S, \mathbb{D}_T)$ denote the population $\mathcal{H}$-divergence between distributions $\mathbb{D}_S$ and $\mathbb{D}_T$, and $\hat{d}_{\mathcal{H}}(\mathcal{D}_S, \mathcal{D}_T)$ be its empirical estimation from datasets $\mathcal{D}_S$ and $\mathcal{D}_T$. 
For any $\eta \in (0,1)$, with probability at least $1-\eta$, }
\begin{equation}
    d_{\mathcal{H}}\left(\mathbb{D}_S, \mathbb{D}_T\right) \leq \hat{d}_{\mathcal{H}}\left(\mathcal{D}_S, \mathcal{D}_T\right)  + 
    4  \max \{ \sqrt{\frac{d \ln (2N_S)+\ln (\frac{2}{\eta})}{N_S}}, 
    \sqrt{\frac{d \ln (2N_T)+\ln (\frac{2}{\eta})}{N_T} } \} , 
\end{equation}
\textit{where $N_S$ is the number of instances in $\mathcal{D}_S$ and $N_T$ is the number of instances in $\mathcal{D}_T$. 
}

\textit{Proof. 
According to \citeec{kifer2004detecting2}, we have }
\begin{equation}
    \begin{aligned}
& P \left[\left|
d_{\mathcal{H}}\left(\mathbb{D}_S, \mathbb{D}_T\right)- \hat{d}_{\mathcal{H}}\left(\mathcal{D}_S,\mathcal{D}_T\right)\right|>\epsilon\right] \leq \left(2 N_S\right)^d e^{-N_S \epsilon^2 / 16}+\left(2 N_T\right)^d e^{-N_T \epsilon^2 / 16}. 
\end{aligned}
\end{equation}

\textit{By replacing $\left(2 N_S\right)^d e^{-N_S \epsilon^2 / 16}+\left(2 N_T\right)^d e^{-N_T \epsilon^2 / 16}$ with $\eta$, we get }
\begin{equation}
    \eta \leq 2 \max \{ \left(2 N_S\right)^d e^{-N_S \epsilon^2 / 16},  
    \left(2 N_T\right)^d e^{-N_T \epsilon^2 / 16} \},
\end{equation}

\textit{which means that $\eta \le 2 \left(2 N_S\right)^d e^{-N_S \epsilon^2 / 16}$ as well as $ \eta \le 2 \left( 2 N_T\right)^d e^{-N_S \epsilon^2 / 16}$. In other words, we have $\epsilon \le 4\sqrt{(d \ln (2N_S)+\ln (2/\eta)) / N_S}$ as well as $\epsilon \le 4\sqrt{(d \ln (2N_T)+\ln (2/\eta))/N_T}$.
Because increasing $\epsilon$ will decrease the probability $P \left[\left|
d_{\mathcal{H}}\left(\mathbb{D}_S, \mathbb{D}_T\right)- \hat{d}_{\mathcal{H}}\left(\mathcal{D}_S,\mathcal{D}_T\right)\right|>\epsilon\right]$, for}
\begin{equation}
    \epsilon_{\max} =  4  \max \{ \sqrt{\frac{d \ln (2N_S)+\ln (\frac{2}{\eta})}{N_S}}, 
    \sqrt{\frac{d \ln (2N_T)+\ln (\frac{2}{\eta})}{N_T} } \}, 
\end{equation}
\textit{we have }
\begin{equation}
    \begin{aligned}
\eta  
& \geq P\left[\left|
d_{\mathcal{H}}\left(\mathbb{D}_S, \mathbb{D}_T\right)- \hat{d}_{\mathcal{H}}\left(\mathcal{D}_S,\mathcal{D}_T\right)\right|>\epsilon\right] \\
& \geq P\left[\left|
d_{\mathcal{H}}\left(\mathbb{D}_S, \mathbb{D}_T\right)- \hat{d}_{\mathcal{H}}\left(\mathcal{D}_S,\mathcal{D}_T\right)\right|>\epsilon_{\max} \right]
\end{aligned}
\end{equation}
\textit{which concludes the proof of Lemma 2.
If $N_S = N_T$, Lemma 2 will reduce to Lemma 1 in \citeec{ben2010theory2}}.




\textbf{Definition A.2. (The ideal hypothesis)}. The ideal hypothesis $h^*$ is defined as the hypothesis that minimizes the summation of the expected differences between $h\in \mathcal{H}$ and the target domain labeling function $f_T$ in both source and target domains 
\begin{equation}
    h^* = \argmin_{h\in\mathcal{H}} \epsilon_S(h,f_T) + \epsilon_T(h,f_T),
\end{equation}
where $\epsilon_S(h,f_T)= E_{\textbf{x}\sim \mathbb{D}_S} [|h(\textbf{x})-f_T(\textbf{x})| ]$ and $\epsilon_T(h,f_T)= E_{\textbf{x}\sim \mathbb{D}_T} [|h(\textbf{x})-f_T(\textbf{x})| ]$.

The error incurred by the ideal hypothesis is $\lambda = \epsilon_S(h^*,f_T) + \epsilon_T(h^*,f_T)$.
Now, we provide the proof of Theorem~\ref{theorem:single} utilizing the lemmas established above.

\textbf{Theorem 1 }  
\textit{For a source domain instance with features $\textbf{x}_i^S$, let $c(\textbf{x}_i^{S})$ be the features of its nearest target domain instance, where $i=1,2,\dots,N_S$ and the distance between a pair of instances are measured over their feature space. If $f_T$ is $L$-Lipschitz continuous, then for any $\eta\in (0,1)$, with probability at least $(1-\eta)^2$, }
\begin{equation} 
\begin{aligned}
\epsilon_T(h) \leq \epsilon_S(h) + \hat{d}_{\mathcal{H}}(\mathcal{D}_S,\mathcal{D}_T) + \frac{1}{N_S} \sum_{i=1}^{N_S} |f_S(\textbf{x}_i^S) - f_T(c(\textbf{x}_i^{S}))| + \frac{L}{N_S} \sum_{i=1}^{N_S} ||\textbf{x}_i^S - c(\textbf{x}_i^{S})||  + C_1, 
\end{aligned}
\end{equation} 
\textit{where $L$ is the Lipschitz constraint constant, $\hat{d}_{\mathcal{H}}(\mathcal{D}_S,\mathcal{D}_T)$ denotes the empirical $\mathcal{H}$-divergence between distributions $\mathbb{D}_S$ and $\mathbb{D}_T$, 
$||\textbf{x}_i^S - c(\textbf{x}_i^{S})||$ represents the distance between $\textbf{x}_i^S$ and $c(\textbf{x}_i^{S})$, 
and $C_1 =\lambda + \sqrt{-\frac{\ln(\frac{\eta}{2})}{2 N_S} } + 2 \max \{ \sqrt{\frac{d \ln (2N_S)+\ln (\frac{2}{\eta})}{N_S}},  \sqrt{\frac{d \ln (2N_T)+\ln (\frac{2}{\eta})}{N_T} } \} $. }

\textit{Proof. With probability at least $(1-\eta)^2$, } 
\begin{equation}
\begin{aligned}
\epsilon_T(h) 
& \leq \epsilon_T\left(h^*\right)+\epsilon_S(h)+\epsilon_S\left(h^*\right)+\frac{1}{2} d_{\mathcal{H} }\left(\mathbb{D}_S, \mathbb{D}_T \right)  \\
& \leq \epsilon_T\left(h^*\right)+\epsilon_S(h)+\epsilon_S\left(h^*,f_T\right) + \epsilon_S\left(f_S,f_T\right) + \frac{1}{2} d_{\mathcal{H} }\left(\mathbb{D}_S, \mathbb{D}_T\right) \\
& =\epsilon_S(h)+\frac{1}{2} d_{\mathcal{H} }\left( \mathbb{D}_S,\mathbb{D}_T \right) + \epsilon_S\left(f_S,f_T\right)+ \lambda \\
& \leq \epsilon_S(h) +\frac{1}{2} \hat{d}_{\mathcal{H} }(\mathcal{D}_S,\mathcal{D}_T) + 
\frac{1}{N_S} \sum_{i=1}^{N_S} |f_S(\textbf{x}_i^S) - f_T(\textbf{x}_i^S) | + C_1 \\
&= \epsilon_S(h) +\frac{1}{2} \hat{d}_{\mathcal{H} }(\mathcal{D}_S,\mathcal{D}_T) + \frac{1}{N_S} \sum_{i=1}^{N_S} |f_S(\textbf{x}_i^S) + f_T(c(\textbf{x}_i^{S})) - f_T(c(\textbf{x}_i^{S})) - f_T(\textbf{x}_i^S) | + C_1  \\ 
&\leq \epsilon_S(h) +\frac{1}{2} \hat{d}_{\mathcal{H} }(\mathcal{D}_S,\mathcal{D}_T) + \frac{1}{N_S} \sum_{i=1}^{N_S} |f_S(\textbf{x}_i^S) - f_T(c(\textbf{x}_i^{S}))| + 
\frac{1}{N_S} \sum_{i=1}^{N_S} |f_T(c(\textbf{x}_i^{S})) - f_T(\textbf{x}_i^S) | + C_1 \\ 
&\leq \epsilon_S(h) +\frac{1}{2} \hat{d}_{\mathcal{H} }(\mathcal{D}_S,\mathcal{D}_T) + \frac{1}{N_S} \sum_{i=1}^{N_S} |f_S(\textbf{x}_i^S) - f_T(c(\textbf{x}_i^{S}))| + \frac{L}{N_S} \sum_{i=1}^{N_S} ||\textbf{x}_i^S- c(\textbf{x}_i^{S})|| + C_1. 
\end{aligned}
\end{equation}

\textit{The first step is based on the proof of Theorem 2 in \citeec{ben2010theory2}. The second step is followed by triangle inequality. The third step utilizes the definition $\lambda = \epsilon_S(h^*,f_T) + \epsilon_T(h^*,f_T)$ and the definition $\epsilon_T\left(h^*\right)=\epsilon_T(h^*,f_T)$.
The fourth step employs Lemma 1 and Lemma 2 to respectively expand the terms $\epsilon_S\left(f_S,f_T\right)$ and $d_{\mathcal{H} }(\mathbb{D}_S,\mathbb{D}_T)$ while collecting the constants into $C_1$. Because the inequalities given by Lemma 1 and Lemma 2 hold both with at least probability $1-\eta$, the inequality at the second step holds with at least probability $(1-\eta)^2$ under the independence assumption.
The fifth step adds and subtracts $f_T(c(\textbf{x}_i^{S}))$. The sixth step utilizes triangle inequality. The last step is from the assumption of $L$-Lipschitz continuous for $f_T$. }

By extending Theorem 1 to the case of multiple source domains, we have the following proposition. 

\textbf{Proposition 1 } 
\textit{For a source domain instance with features $\textbf{x}_i^{S_j}$ in $ \mathcal{D}_{S_j}$, let $c(\textbf{x}_i^{S_j})$ be the features of its nearest instance in $\mathcal{D}_T$, where $j=1,2,\dots,k$ and $i=1,2,\dots,N_{S_j}$. If $f_T$ is $L$-Lipschitz continuous, then for any $\eta\in(0,1)$, with probability at least $(1-\eta)^{2k} $, } 
\begin{equation}
\epsilon_T(h) \leq \frac{1}{k} \sum_{j=1}^k \{ \epsilon_{S_j}(h) + \frac{1}{2} \hat{d}_{\mathcal{H} }(\mathcal{D}_{S_j},\mathcal{D}_T)  + \frac{1}{N_{S_j}} \sum_{i=1}^{N_{S_j}} |f_{S_j}(\textbf{x}_i^{S_j}) - f_T(\textbf{x}_i^{{S_j},T})| + \frac{L}{N_{S_j}} \sum_{i=1}^{N_{S_j}} ||\textbf{x}_i^{S_j}-\textbf{x}_i^{{S_j},T}||  +C_{j} \}, 
\end{equation} 
\textit{where $C_j=\lambda_j + \sqrt{-\frac{\ln(\frac{\eta}{2})}{2 N_{S_j}}}+ 2 \max \{ \sqrt{\frac{d \ln (2N_{S_j})+\ln (\frac{2}{\eta})}{N_{S_j}}}, \sqrt{\frac{d \ln (2N_T)+\ln (\frac{2}{\eta})}{N_T} } \}$, $\lambda_j$ is the error incurred by the ideal hypothesis $\lambda_j = \epsilon_{S_j}(h^*,f_T) + \epsilon_T(h^*,f_T)$, and $L$ is the constant for the Lipschitz condition. }

\textit{Proof. By Theorem 1, for any source domain $j=1,2,\cdots,k$, the following inequality holds with probability at least $(1-\eta)^2$ for any $\eta\in (0,1)$,}
\begin{equation} 
\begin{aligned}
\epsilon_T(h) \leq \epsilon_{S_j}(h) + \hat{d}_{\mathcal{H}}(\mathcal{D}_{S_j},\mathcal{D}_T) + \frac{1}{N_{S_j}} \sum_{i=1}^{N_{S_j}} |f_{S_j}(\textbf{x}_i^{S_j}) - f_T(c(\textbf{x}_i^{S_j}))| + \frac{L}{N_{S_j}} \sum_{i=1}^{N_{S_j}} ||\textbf{x}_i^{S_j} - c(\textbf{x}_i^{S_j})||  + C_j. 
\end{aligned}
\end{equation} 
\textit{ Proposition 1 follows immediately by summing on each side of inequality (A17) over all source domains.
}

\section{Implementation Details of Benchmark Methods} \label{appendix:benchmark}

\textbf{MMD. } Maximum Mean Discrepancy (MMD) is a metric to measure the discrepancy between distributions in a reproducing kernel Hilbert space (RKHS) \citepec{borgwardt2006integrating2}. Specifically, a kernel function $k$ is associated with a feature map $\phi$, i.e., $k(\textbf{x}_S,\textbf{x}_T)=<\phi(\textbf{x}_S), \phi(\textbf{x}_T)>$, where $\textbf{x}_S$ and $\textbf{x}_T$ are source and target distributions. 
Let $\mathcal{H}$ be an RKHS, and MMD is defined as the discrepancy between the mean embeddings of source and target distributions in $\mathcal{H}$:
\begin{equation}
    \text{MMD} = d(p_S,p_T)^2 = ||\frac{1}{N_S}\phi(\textbf{x}_S) - \frac{1}{N_T}\phi(\textbf{x}_T) ||_{\mathcal{H}}^2. 
\end{equation}
As reviewed in Section~\ref{sec:rw:cross}, MMD is an explicit metric that measures the discrepancy between source and target domains in terms of their feature distributions. 
In our implementation, we utilized the Gaussian kernel $k(\textbf{x}_S,\textbf{x}_T) = e^{||\textbf{x}_S-\textbf{x}_T|| / \gamma}$ and set $\gamma=0.5$.

\textbf{CANMD. } Contrastive Adaptation Network for Early Misinformation Detection (CANMD) is a domain adaptation method devised for misinformation detection \citepec{yue2022contrastive2}. 
This method assigns pseudo labels to target domain instances. Let $D_{ij}$ denote the multi-kernel MMD discrepancy between the covariate distribution of source domain instances with label $i$ and that of target domain instances with pseudo label $j$, where $i,j\in\{0,1\}$. To minimize the divergence between source and target domains, the loss is defined as 
\begin{equation}
    \mathcal{L} = D_{00} + D_{11} - \frac{1}{2} (D_{01}+D_{10}).
\end{equation}
Following \citeec{yue2022contrastive2}, we implemented the multi-kernel MMD 
using 5 Gaussian kernels with $k(\textbf{x}_S,\textbf{x}_T) = e^{||\textbf{x}_S-\textbf{x}_T|| / \gamma}$ and $\gamma=2^k,$ where the value of $k$ for each Gaussian kernel was set to -3, -2, -1, 0, and 1, respectively. The Gaussian kernels were combined with equal weights. 

\textbf{DANN. } The Domain Adversarial Neural Networks (DANN) method has two classifiers: a label classifier and a domain classifier \citepec{ganin2016domain2,li2021multi2}. The former is learned to classify each piece of source or target domain information as true or false. The latter is used to learn domain-invariant features. The hyperparameter $\lambda$ of DANN controls the relative weight between its domain and label classifiers. In our evaluation, we performed a grid search for the value of $\lambda$ from 1 to 10, with an interval of 1. We found that $\lambda=5$ yielded the best performance for DANN and chose $\lambda=5$ accordingly. 


\textbf{MK-MMD. } Multi-kernel MMD (MK-MMD) is an extension of single-kernel MMD \citepec{long2015learning2}. Instead of using a single kernel function, MK-MMD employs a convex combination of multiple kernel functions. In our implementation, we used Gaussian kernels $k(\textbf{x}_S,\textbf{x}_T) = e^{||\textbf{x}_S-\textbf{x}_T|| / \gamma}$ and $\gamma=2^k,$ where the value of $k$ for each Gaussian kernel was set to -3, -2, -1, 0, and 1, respectively. The Gaussian kernels were combined with equal weights. 


\textbf{JAN. } Joint Adaptation Networks (JAN) measures the discrepancy between source and target domains at each layer of its classifier using MK-MMD \citepec{long2017deep2}. It then combines these discrepancies and minimizes the aggregated discrepancy. The implementation of MK-MMD in JAN is the same as the implementation of MK-MMD described above.


\textbf{DALN. } Discriminator-free Adversarial Learning Network (DALN) utilizes the Nuclear-norm Wasserstein Discrepancy (NWD) to measure the covariate shift \citepec{chen2022reusing2}. To compute NWD, \citeec{chen2022reusing2} devise a domain classifier that predicts the probability of an instance belonging to the source domain. NWD is then defined as the difference between the average probability of source domain instances and that of target domain instances. The DALN method minimizes the covariate shift by minimizing NWD. In addition, the DALN method has a label classifier that predicts the label (true or false) for each instance. The loss of the method is a summation of NWD and the label classification loss. 

\textbf{MCD. } The Maximum Classifier Discrepancy (MCD) method consists of three steps \citepec{saito2018maximum2}. In the first step, it trains two classifiers using labeled source domain instances. 
Next, it maximizes the discrepancy between two classifiers when applied to target domain instances. In the third step, it adjusts its feature extractor to minimize such discrepancy. 
MCD can be viewed as a generative adversarial network (GAN), with the feature extractor serving as the generator and the two classifiers functioning as the discriminator. 


\textbf{SDAT. } Smooth Domain Adversarial Training (SDAT) is an implicit domain adaptation method based on DANN \citepec{rangwani2022closer2}. 
Inspired by the observation that the label classifier of DANN often falls into sharp minima, 
\citeec{rangwani2022closer2} devise a smooth training loss based on the Sharpness Aware Minimization (SAM) proposed by \citeec{foret2020sharpness2}. 
In our implementation, we set the neighborhood size $\rho$ in SAM to 0.05 by following \citepec{foret2020sharpness2}.




\textbf{REAL-FND. } REinforced Adaptive Learning Fake News Detection (REAL-FND) first pretrains a domain classifier and a label classifier \citepec{mosallanezhad2022domain_ec}. Then, reinforcement learning is applied on content embeddings to maximize the confidence of the label classifier and adversary confidence of the domain classifier. Following \citetec{mosallanezhad2022domain_ec}, the reinforcement learning agent employs a simple feed-forward network with 2 layers. Each episode of the agent contains 20 actions to adjust the embedding.

\section{Robustness Analysis}\label{appendix:robustness}

\subsection{The Second Infodemic Dataset: Detecting Misinformation in Chinese News}

\label{appendix:chinese}

We conducted an additional empirical evaluation using a public dataset of Chinese news \citepec{nan2021mdfend2}, which has been widely adopted in recent misinformation studies \citepec[e.g.,][]{wu2023mfir2,zhu_memory-guided_2023_2,peng2024not2}. The dataset contains true and fake news articles in nine different domains. The fake news articles in the dataset were labeled by the Weibo Community Management Center, whereas the true news articles were verified by NewsVerify, a platform dedicated to evaluating the credibility of content on Weibo. We employed news articles in three of these domains: entertainment, politics, and health, consistent with the setting of the empirical study with English news datasets. Not all news articles in the health domain are related to infodemic. We specifically chose news articles from this domain that contain words associated with the COVID pandemic, such as COVID, COVID test, or pandemic. Table~\ref{tab:datastat_cn} lists the number of true and fake news articles in each domain. User comments accompany 53.22\% of COVID news, 84.27\% of politics news, and 94.64\% of entertainment news. The average comment length is 16.4 words in the COVID domain, 14.1 words in the entertainment domain, and 18.8 words in the politics domain. 

\begin{table*}[htbp]
\centering
\caption{Summary Statistics of Chinese News Dataset (Three Domains). }
\def\arraystretch{1.3}\begin{tabular}{lccc}
\toprule 
Domain & Number of true news & Number of fake news \\
\midrule
COVID (Infodemic / Target) & 201 & 124  \\
Entertainment (Source) & 1000 & 440   \\ 
Politics (Source) & 306 & 546 \\
\bottomrule 
\end{tabular}%
\label{tab:datastat_cn}
\end{table*}%

We followed the evaluation procedure  described in Section \ref{sec:setup}, except that Chinese words in the dataset were embedded using the pretrained BERT-wwm model \citepec{cui2021pre2}.
Table~\ref{tab:improvement_cn} compares the performance between our method and each benchmark method in evaluation scenario 1. As reported, our method significantly outperforms each benchmark, improving precision by a range of 3.53\% to 15.22\%, recall by a range of 3.88\% to 25.33\%, F1-score by a range of 3.33\% to 25.12\%, and F2-score by a range of 3.87\% to 22.50\%.

\begin{table*}[htbp]
\centering
\caption{Performance Comparison between Our Method and Benchmark Methods using the Chinese News Dataset}
\def\arraystretch{1.2}
\begin{tabular}{ p{2cm} p{2cm}<{\centering} p{2cm}<{\centering} p{2cm}<{\centering} p{2cm}<{\centering} } 
\toprule 
Method & Precision & Recall & F1-score & F2-score \\
\midrule
\multirow{2}{*}{DACA}  & \textbf{0.822} & \textbf{0.836} & \textbf{0.824} & \textbf{0.833} \\ \addlinespace[-1ex]
& (0.010) & (0.004) & (0.010) & (0.017) \\ 
\multirow{2}{*}{MMD} & 0.794$^{**}$ & 0.804$^{**}$ & 0.797$^{**}$ & 0.802$^{**}$ \\ \addlinespace[-1ex]
& (0.001) & (0.006) & (0.001) & (0.013) \\
\multirow{2}{*}{CANMD} & 0.788$^{**}$ & 0.792$^{**}$ & 0.789$^{**}$ & 0.791$^{**}$ \\ \addlinespace[-1ex]
& (0.011) & (0.013) & (0.012) & (0.024) \\
\multirow{2}{*}{DANN} & 0.716$^{**}$ & 0.713$^{**}$ & 0.709$^{**}$ & 0.713$^{**}$ \\ \addlinespace[-1ex]
& (0.007) & (0.002) & (0.001) & (0.031) \\
\multirow{2}{*}{MK-MMD} & 0.778$^{**}$ & 0.786$^{**}$ & 0.780$^{**}$ & 0.784$^{**}$\\ \addlinespace[-1ex]
& (0.010) & (0.006) & (0.009) & (0.017) \\
\multirow{2}{*}{JAN}  & 0.717$^{**}$ & 0.723$^{**}$ & 0.718$^{**}$ & 0.722$^{**}$ \\ \addlinespace[-1ex]
& (0.023) & (0.014) & (0.023) & (0.034) \\
\multirow{2}{*}{DALN}  & 0.794$^{**}$ & 0.764$^{**}$ & 0.763$^{**}$ & 0.769$^{**}$ \\ \addlinespace[-1ex]
& (0.007) & (0.013) & (0.003) & (0.034) \\
\multirow{2}{*}{MCD}  & 0.739$^{**}$ & 0.667$^{**}$ & 0.658$^{**}$ & 0.680$^{**}$ \\ \addlinespace[-1ex]
& (0.044) & (0.074) & (0.082) & (0.056)   \\ 
\multirow{2}{*}{SDAT}  & 0.771$^{**}$ & 0.777$^{**}$ & 0.772$^{**}$ & 0.776$^{**}$ \\ \addlinespace[-1ex]
& (0.012) & (0.002) & (0.008) & (0.017)  \\
\bottomrule 
\end{tabular}%
\begin{tablenotes}
\centering\item[*] Note: Significance levels are denoted by * and ** for 0.05 and 0.01, respectively. 
\end{tablenotes}
\label{tab:improvement_cn}
\end{table*}

\subsection{The Third Infodemic Dataset: Detecting Misinformation in Social Media Posts}
\label{appendix:constraint}

We employed the Constraint dataset \citepec{patwa2021fighting_ec} as the third COVID infodemic dataset.
This dataset comprises 10,700 social media posts related to COVID-19, with an average length of 27.05 English words per post. Among these posts, 5,100 posts are labeled as fake while 5,600 as true. The fake posts are collected from fact-checking websites whereas the true posts are gathered from verified Twitter accounts \citepec{patwa2021fighting_ec}. 
Considering that social media posts are typically much shorter than news articles, we selected two source domain datasets --- PHEME \citepec{zubiaga_learning_2016_ec} and FEVER \citepec{thorne2018fever_ec} --- whose document lengths are comparable to that of the Constraint dataset. The PHEME dataset is a collection of Twitter posts related to breaking news events, including 48,619 true and 13,824 fake posts, annotated by a team of journalists. The average post length is 21.5 words. 
The FEVER dataset consists of 80,035 true claims and 29,775 false claims, all human-labeled, with an average claim length of 9.4 words.\footnotemark\footnotetext{We used a preprocessed version of the PHEME dataset, which can be accessed at \url{https://www.kaggle.com/datasets/nicolemichelle/pheme-dataset-for-rumour-detection/data}.}
We followed the experimental procedure  described in Section \ref{sec:setup} to evaluate the performance of our method and the benchmark methods. Table~\ref{tab:ec:claim} presents the average precision, recall, F1-score, and F2-score for each method, along with standard deviations (in parentheses), across 25 experiments. As reported, our DACA method significantly outperforms each benchmark method across all evaluation metrics. 
For example, DACA achieves an average recall of 0.758, surpassing that of the best-performing benchmark method (MCD) by 6.61\%. 

\begin{table*}[htbp]
\centering
\caption{Performance Comparison between Our Method and Benchmark Methods using the Constraint Dataset.}
\def\arraystretch{1.1}
\begin{tabular} { p{2cm} p{2.5cm}<{\centering} p{2.5cm}<{\centering} p{2.5cm}<{\centering} p{2.5cm}<{\centering}} 
\toprule 
Method & Precision & Recall & F1-score & F2-score\\
\midrule
\multirow{2}{*}{DACA} & \textbf{0.768} & \textbf{0.758} & \textbf{0.757} & \textbf{0.760} \\ \addlinespace[-1ex]
& (0.041) & (0.042) & (0.044) & (0.042) \\ 
\multirow{2}{*}{MMD}  & 0.620$^{**}$ & 0.606$^{**}$ & 0.597$^{**}$ & 0.608$^{**}$ \\ \addlinespace[-1ex]
& (0.042) & (0.029) & (0.036) & (0.030) \\
\multirow{2}{*}{CANMD} & 0.710$^{**}$ & 0.700$^{**}$ & 0.698$^{**}$ & 0.703$^{**}$ \\ \addlinespace[-1ex]
& (0.039) & (0.037) & (0.038) & (0.037) \\
\multirow{2}{*}{DANN} & 0.712$^{**}$ & 0.704$^{**}$ & 0.702$^{**}$ & 0.705$^{**}$ \\ \addlinespace[-1ex]
& (0.046) & (0.045) & (0.047) & (0.045) \\

\multirow{2}{*}{MK-MMD} & 0.608$^{**}$ & 0.598$^{**}$ & 0.590$^{**}$ & 0.600$^{**}$ \\ \addlinespace[-1ex]
& (0.041) & (0.038) & (0.041) & (0.038) \\
\multirow{2}{*}{JAN} & 0.644$^{**}$ & 0.631$^{**}$ & 0.622$^{**}$ & 0.633$^{**}$ \\ \addlinespace[-1ex]
& (0.047) & (0.038) & (0.040) & (0.063) \\
\multirow{2}{*}{DALN} & 0.681$^{**}$ & 0.664$^{**}$ & 0.657$^{**}$ & 0.667$^{**}$ \\ \addlinespace[-1ex]
& (0.041) & (0.040) & (0.044) & (0.040) \\

\multirow{2}{*}{MCD} & 0.734$^{*}$ & 0.711$^{**}$ & 0.704$^{**}$ & 0.715$^{**}$ \\ \addlinespace[-1ex]
& (0.049) & (0.061) & (0.078) & (0.058) \\ 
\multirow{2}{*}{SDAT} & 0.663$^{**}$ & 0.649$^{**}$ & 0.640$^{**}$ & 0.652$^{**}$ \\ \addlinespace[-1ex]
& (0.040) & (0.046) & (0.057) & (0.045) \\
\bottomrule 
\end{tabular}%
\begin{tablenotes}
\centering\item[*] Note: Significance levels are denoted by * and ** for 0.05 and 0.01, respectively. 
\end{tablenotes}
\label{tab:ec:claim}
\end{table*}

\subsection{Alternative Target Domain: Detecting Financial Misinformation}
\label{appendix:finance}

We consider the diffusion of true and false information in the finance domain and target the task of detecting financial misinformation, which is an emerging but increasingly important research area with substantial implications for market stability and investor protection \citepec{clarke_fake_2021_ec, rangapur_investigating_2023_ec, liu_fmdllama_2025_ec}. 
By applying our proposed method to this domain, we aim to show that the concept alignment mechanism can effectively extract domain-invariant misinformation signals for other contexts as well.

In this evaluation, we utilized the Chinese news dataset \citepec{nan2021mdfend2}, details of which are summarized in Appendix \ref{appendix:chinese}. We employed the politics and entertainment domains in this dataset as the source domains, and the finance domain as the target domain.
Table~\ref{tab:ec:datastat_cn} lists the number of true and fake news in each domain. 
We followed the experimental procedure  described in Section \ref{sec:setup} to evaluate the performance of our method and the benchmark methods.
Table~\ref{tab:ec:cn_finance} presents the average precision, recall, F1-score, and F2-score for each method, along with standard deviations (in parentheses), across 25 experiments. As reported, our DACA method significantly outperforms each benchmark method across all metrics.
This additional evaluation demonstrates the effectiveness of our method in a context beyond COVID.

\begin{table*}[htbp]
\centering
\caption{Summary Statistics of Chinese News Dataset (Financial Misinformation Detection). }
\def\arraystretch{1.3}\begin{tabular}{lccc}
\toprule 
Domain & Number of true news & Number of fake news \\
\midrule
Finance (Target) & 959 & 362  \\
Entertainment (Source) & 1000 & 440   \\ 
Politics (Source) & 306 & 546 \\
\bottomrule 
\end{tabular}%
\label{tab:ec:datastat_cn}
\end{table*}%

\begin{table*}[htbp]
\centering
\caption{Performance Comparison between Our Method and Benchmark Methods on Financial Misinformation Detection.}
\def\arraystretch{1.1}
\begin{tabular} { p{2cm} p{2.5cm}<{\centering} p{2.5cm}<{\centering} p{2.5cm}<{\centering} p{2.5cm}<{\centering}} 
\toprule 
Method & Precision & Recall & F1-score & F2-score\\
\midrule
\multirow{2}{*}{DACA} & \textbf{0.804} & \textbf{0.823} & \textbf{0.810} & \textbf{0.819} \\ \addlinespace[-1ex]
& (0.034) & (0.027) & (0.030) & (0.027) \\ 
\multirow{2}{*}{MMD}  & 0.770$^{**}$ & 0.786$^{**}$ & 0.775$^{**}$ & 0.783$^{*}$ \\ \addlinespace[-1ex]
& (0.034) & (0.031) & (0.033) & (0.031) \\
\multirow{2}{*}{CANMD} & 0.745$^{**}$ & 0.786$^{**}$ & 0.755$^{**}$ & 0.777$^{**}$ \\ \addlinespace[-1ex]
& (0.038) & (0.035) & (0.039) & (0.036) \\
\multirow{2}{*}{DANN} & 0.714$^{**}$ & 0.694$^{**}$ & 0.690$^{**}$ & 0.697$^{**}$ \\ \addlinespace[-1ex]
& (0.045) & (0.065) & (0.057) & (0.058) \\

\multirow{2}{*}{MK-MMD} & 0.762$^{**}$ & 0.782$^{**}$ & 0.769$^{**}$ & 0.778$^{**}$ \\ \addlinespace[-1ex]
& (0.042) & (0.037) & (0.041) & (0.038) \\
\multirow{2}{*}{JAN} & 0.743$^{**}$ & 0.776$^{**}$ & 0.752$^{**}$ & 0.769$^{**}$ \\ \addlinespace[-1ex]
& (0.041) & (0.041) & (0.044) & (0.040) \\
\multirow{2}{*}{DALN} & 0.774$^{*}$ & 0.761$^{**}$ & 0.762$^{**}$ & 0.763$^{**}$ \\ \addlinespace[-1ex]
& (0.034) & (0.036) & (0.034) & (0.032) \\

\multirow{2}{*}{MCD} & 0.740$^{**}$ & 0.786$^{**}$ & 0.737$^{**}$ & 0.777$^{**}$ \\ \addlinespace[-1ex]
& (0.025) & (0.021) & (0.040) & (0.022) \\ 
\multirow{2}{*}{SDAT} & 0.689$^{**}$ & 0.708$^{**}$ & 0.700$^{**}$ & 0.704$^{**}$ \\ \addlinespace[-1ex]
& (0.044) & (0.041) & (0.059) & (0.041) \\
\bottomrule 
\end{tabular}%
\begin{tablenotes}
\centering\item[*] Note: Significance levels are denoted by * and ** for 0.05 and 0.01, respectively. 
\end{tablenotes}
\label{tab:ec:cn_finance}
\end{table*}

\subsection{Beyond Binary Classification: Forecasting the Number of News Comments}
\label{appendix:regression}

We expand the scope of our study by extending our proposed method to a new task. Specifically, the new task is to predict the number of comments a piece of information (e.g., a news article) will receive. Understandably, this new task is fundamentally different from the original misinformation detection task, as the former is a regression problem while the latter is a binary classification task. Moreover, the number of comments a piece of information receives is a natural proxy for its diffusion and virality, as a high (or low) comment count typically indicates broad (or limited) diffusion and strong (or weak) audience engagement. 
Solving both tasks in tandem enables more refined infodemic management. While the original task focuses on detecting misinformation, the new task assesses the potential impact of a piece of information by predicting its comment count. Solving these two tasks together helps platforms prioritize their resources toward combating misinformation with high virality potential, as such content more likely leads to severe consequences.

We now formally define the new task. Let $\mathcal{D}_{S_j}$ denote a dataset of $N_{S_j}$ pieces of labeled information in the source domain $S_j$, $j=1,2,\dots,k$. Each piece of labeled information in $\mathcal{D}_{S_j}$ is represented as $(\textbf{x}_i^{S_j},y_i^{S_j})$, where $\textbf{x}_i^{S_j}$ denotes features extracted from the information, label $y_i^{S_j}$ represents the number of comments associated with the information, and $i=1,2,\dots,N_{S_j}$.  
Let $\mathcal{D}_{T}$ denote a dataset of $N_{T}$ pieces of unlabeled information in the infodemic domain $T$. Each piece of unlabeled information in $\mathcal{D}_{T}$ is represented as $\textbf{x}_i^T$, i.e., features extracted from the information, $i=1,2,\dots,N_{T}$. The new task is defined as follows:

Given a dataset $\mathcal{D}_T$ of $N_T$ pieces of unlabeled information in the infodemic domain and $k$ datasets of labeled information in various source domains, $\mathcal{D}_{S_1}, \mathcal{D}_{S_2},\dots, \mathcal{D}_{S_k}$, where $\mathcal{D}_{S_j}$ consists of $N_{S_j}$ pieces of labeled information, $j=1,2,\dots,k$, the objective is to learn a model from the data that predicts the number of comments each piece of information in $\mathcal{D}_T$ will receive.

We modified our DACA method for the new task. First, we replaced the classification loss with the mean squared error loss to accommodate the training of a regression task. Second, to learn the similarity function, we calculated the median number of comments received among all labeled source domain information and divided them into two groups: one containing information receiving above-median comment counts and the other with below-median counts. Pairs from the same group were labeled as positive, while pairs from different groups were labeled as negative. 
For each benchmark method, we also replaced the classification loss with the mean square error loss.\footnotemark\footnotetext{We excluded MCD and SDAT from the benchmark list because adapting these methods to a regression setting is nontrivial. Specifically, MCD relies on a distance measure that is inherently defined within a classification framework, while SDAT incorporates a smoothing loss that lacks a direct counterpart for regression tasks.}
We measured the performance of each method on the new task using standard metrics for regression models \citepec{bandari_pulse_2012_ec}: the 
root mean square error (RMSE) and the mean absolute error (MAE).
Let $y_i^{T}$ denote the actual number of comments received by a piece of infodemic information, and let $\hat{y}_i^{T}$ be the corresponding predicted number, where $i = 1, 2, \dots, N_T$. 
RMSE is computed as  
\begin{equation}
    \text{RMSE} = \sqrt {\sum_{i=1}^{N_T} \frac{(y_i^{T} - \hat{y}_i^{T})^2}{N_T} },
\end{equation}
and MAE is formulated as 
\begin{equation}
    \text{MAE} = \sum_{i=1}^{N_T} \frac{|y_i^{T} - \hat{y}_i^{T}|}{N_T}.
\end{equation}
Lower RMSE and MAE indicate better predictive performance.

We employed the Chinese news dataset \citepec{nan2021mdfend2}, details of which are summarized in Appendix \ref{appendix:chinese}. We utilized the politics and entertainment domains in this dataset as the source domains, and the COVID domain as the infodemic domain. In the politics domain, the average and maximum number of comments received per piece of information are 96.31 and 1,554, respectively; in the entertainment domain, 170.84 and 1,761; and in the COVID domain, 10.01 and 473.
We followed the experimental procedure  described in Section \ref{sec:setup} to evaluate the performance of our method and the benchmark methods.
Table~\ref{tab:app:severity} presents the average RMSE and MAE for each method, along with standard deviations (in parentheses), across 25 experiments. 
As reported in the table, our method achieves an average RMSE of 18.548 and an average MAE of 7.766, outperforming all benchmark methods. The evaluation results demonstrate the flexibility of our method in addressing infodemic management tasks beyond the binary classification setting. 

\begin{table*}[htbp]
\centering
\caption{Performance Comparison between Our Method and Benchmark Methods on Comment Count Prediction. }
\def\arraystretch{1.2}
\begin{tabular}{ p{2cm} p{2cm}<{\centering} p{2cm}<{\centering}  } 
\toprule 
Method & RMSE$\downarrow$ & MAE$\downarrow$  \\
\midrule
\multirow{2}{*}{DACA}  & \textbf{18.548} & \textbf{7.766} \\ \addlinespace[-1ex]
 & (4.678)  & (1.294) \\ 
\multirow{2}{*}{MMD} &  22.075$^{**}$ & 9.840$^{**}$  \\ \addlinespace[-1ex]
 & (4.917) & (2.744)  \\
\multirow{2}{*}{CANMD} & 21.168$^*$ & 8.699$^*$  \\ \addlinespace[-1ex]
& (4.397) & (1.660) \\
\multirow{2}{*}{DANN} &  21.066$^{*}$ & 10.181$^{**}$ \\ \addlinespace[-1ex]
& (3.771) & (2.010) \\
\multirow{2}{*}{MKMMD} & 28.296$^{**}$ & 13.488$^{**}$ \\ \addlinespace[-1ex]
& (8.011) & (7.123) \\
\multirow{2}{*}{JAN} & 29.899$^{**}$ & 9.097$^{**}$ \\ \addlinespace[-1ex]
& (6.311) & (0.709) \\
\multirow{2}{*}{DALN} & 21.152$^*$ & 9.639$^{**}$ \\ \addlinespace[-1ex]
& (4.694) & (2.461)  \\
\bottomrule 
\end{tabular}%
\label{tab:app:severity}
\begin{tablenotes}
\centering\item[*] Note: Significance levels are denoted by * and ** for 0.05 and 0.01, respectively. 
\end{tablenotes}
\end{table*}

\subsection{Performance Comparison between Our Method and LLM-based Approaches for Misinformation Detection}
\label{appendix:LLM}

We compared the performance of our method with that of Large Language Model-based approaches (LLM-based approaches) for misinformation detection. 
The news articles in our COVID dataset were published in year 2020. Therefore, any LLM trained on data after 2020 may possess background knowledge for understanding and classifying these news (as true or fake). To ensure a fair comparison between LLMs and our method, the reviewer suggests to enforce a cutoff date for LLM training (i.e., restricting the LLMs to training data before 2020).
Re-training a modern LLM from scratch using only data before 2020 would be a solution to enforce the cutoff date. However, this approach is prohibitively costly and beyond the scope of our study.
An alternative is to use earlier LLMs trained on pre-2020 data. However, these models are generally much smaller and less capable. In our experiments, we found that early models such as GPT-2 were unable to follow instructions to determine whether a piece of information is true or fake.\footnotemark\footnotetext{A prompt-based approach (e.g., explicitly instructing an LLM to ``only use knowledge before 2020'') is unreliable. Since the model parameters are fixed after training, there is no mechanism to constrain the model to rely solely on pre-2020 knowledge. As a result, this approach does not provide a faithful enforcement of the cutoff date.}
Given these constraints, we adopted Mistral-7B-Instruct as our base model. Note that the LLM was released after September 2023. Although the training cutoff date for this LLM has not been disclosed, the model may have an inherent look-ahead advantage over our method when detecting fake news in our COVID dataset.

\begin{figure}[htbp]
\FIGURE{\includegraphics[width=1\textwidth]{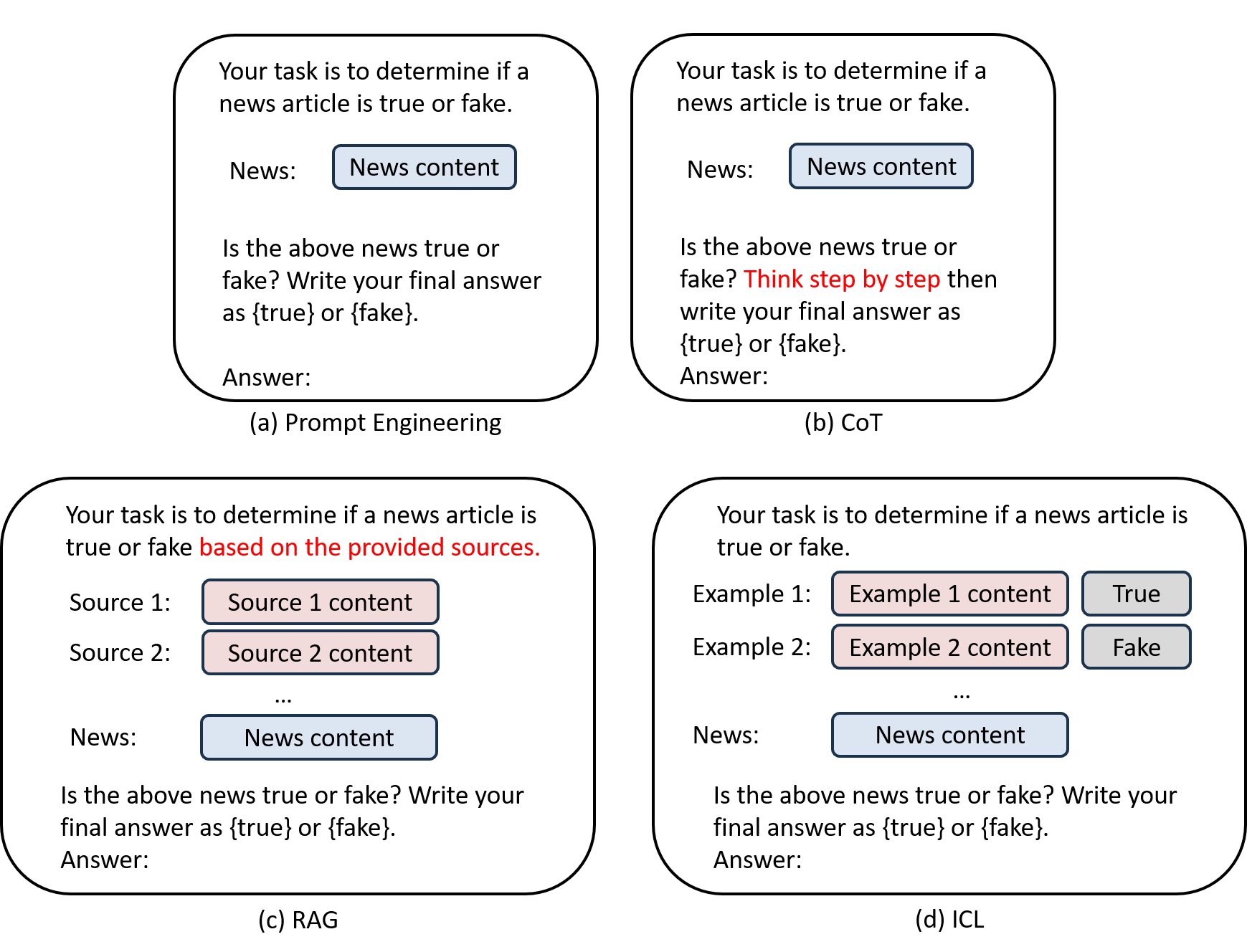}}
{Using LLM for Misinformation Detection.
\label{fig:prompt}}
{Better view in color.}
\end{figure}

We implemented four ways of using Mistral-7B-Instruct for misinformation detection, as illustrated in Figure \ref{fig:prompt}. One way is prompt engineering, where the LLM was directly instructed to judge whether a piece of information is true or fake \citepec{faruk2024evaluating2}. 
Another approach is to leverage chain-of-thought reasoning (CoT). Specifically, the model was asked to think step by step before generating the final answer \citepec{kojima2022large2}.
In addition, we employed retrieval-augmented generation (RAG) to assist the LLM in misinformation detection. Specifically, we randomly sampled 2,000 papers published in 2020 about COVID-19 and coronaviruses from the COVID-19 Open Research Dataset\footnotemark\footnotetext{See \url{https://www.kaggle.com/datasets/allen-institute-for-ai/CORD-19-research-challenge/data}.}. We then utilized the abstracts of these papers as the knowledge base for RAG. 
For the implementation of RAG, we employed a pretrained BERT to compute document embeddings for both the paper abstracts and the news articles. Next, for each news article, the top three most similar abstracts are retrieved from the knowledge base and provided as additional context to help determine its veracity. 
Finally, for in-context learning (ICL), we randomly sampled three news articles from the COVID dataset and attached their true labels. These three labeled news were then presented to the LLM as examples.

\begin{table*}[ht]
\centering
\caption{Performance Comparison between Our Method and LLM-based Approaches for Misinformation Detection.}
\def\arraystretch{1.3}
\begin{tabular}{ p{3.5cm} p{2.5cm} p{2.5cm} p{2.5cm} p{2.5cm} } 
\toprule 
Method & Precision & Recall & F1-score & F2-score \\
\midrule
\multirow{2}{*}{DACA} & \textbf{0.716} & \textbf{0.745} & \textbf{0.719} & \textbf{0.739} \\ \addlinespace[-1ex]
& (0.030) & (0.049) & (0.034) & (0.043) \\
\multirow{2}{*}{Prompt Engineering} & 0.664$^{**}$ & 0.613$^{**}$ & 0.626$^{**}$ & 0.616$^{**}$ \\ 
\addlinespace[-1ex]
& (0.007) & (0.014) & (0.014) & (0.013) \\

\multirow{2}{*}{RAG} & 0.670$^{**}$ & 0.671$^{**}$ & 0.670$^{**}$ & 0.671$^{**}$ \\ 
\addlinespace[-1ex]
& (0.006) & (0.006) & (0.006) & (0.006) \\

\multirow{2}{*}{ICL} & 0.707 & 0.659$^{**}$ & 0.670$^{**}$ & 0.661$^{**}$ \\ 
\addlinespace[-1ex]
& (0.027) & (0.027) & (0.019) & (0.025) \\ 
\bottomrule 
\end{tabular}%
\begin{tablenotes}
\centering\item[*] Note: Significance levels are denoted by * and ** for 0.05 and 0.01, respectively. 
\end{tablenotes}
\label{tab:llm}
\end{table*}

Table \ref{tab:llm} presents the precision, recall, F1-score, and F2-score for each compared method, excluding CoT. For CoT, our empirical evaluation shows that the model labels all news articles as true. A similar pattern is observed when using a reasoning model (DeepSeek-R1-Distill-Llama-8B). This behavior could be attributed to the sycophancy of LLMs \citepec{sharmatowards2}, where models are trained to align with user-provided content, even when it contains misinformation. During the reasoning process, the LLM generates justifications for the claim and consequently outputs a ``true'' label. As reported in the table, our DACA method significantly outperforms all LLM-based approaches in recall, F1-score, and F2-score. It also surpasses these approaches in precision, though its lead over ICL is not statistically significant. Notably, our method achieves this superior performance despite the possible inherent look-ahead advantage of the LLM.

\section{Sensitivity Analysis} \label{appendix:sensitivity}

\subsection{Model Training with Incomplete Source Domain Data}
\label{appendix:incomplete_info}

First, we evaluate the performance of our method when trained using a single source domain. Table~\ref{tab:app:one_source} presents the average performance of DACA (with standard deviations in parentheses) across 25 experiments when trained using both the Politics and Entertainment source domains, as well as the results of DACA (Politics) and DACA (Entertainment) when trained on each domain individually. The results show that using both domains outperforms using either one alone, suggesting that diverse source data enhance model generalizability. Moreover, DACA (Politics) performs better than DACA (Entertainment), indicating that cross-domain learning from the Politics domain contributes more to the performance of our method than learning from the Entertainment domain.

\begin{table*}[htbp]
\centering
\caption{Sensitivity Analysis of Source Domains.}
\def\arraystretch{1.3}
\begin{tabular}{ p{3.5cm} p{2.5cm}<{\centering} p{2.5cm}<{\centering} p{2.5cm}<{\centering} p{2.5cm}<{\centering} } 
\toprule 
 & Precision & Recall & F1-score & F2-score \\
\midrule
\multirow{2}{*}{DACA}  & 0.716 & 0.745 & 0.719 & 0.739 \\ \addlinespace[-1ex]
& (0.030) & (0.049) &  (0.034) & (0.043) \\ 
\multirow{2}{*}{DACA (Politics)}  &  0.685$^{**}$ & 0.727 & 0.692$^{*}$ & 0.718$^{*}$ \\ \addlinespace[-1ex]
& (0.054) & (0.042) & (0.044) & (0.045) \\
\multirow{2}{*}{DACA (Entertainment)}  & 0.665$^{**}$ & 0.697$^{**}$ & 0.672$^{**}$ & 0.690$^{**}$ \\ \addlinespace[-1ex]
& (0.028) & (0.047) & (0.032) & (0.043) \\
\bottomrule 
\end{tabular}%
\begin{tablenotes}
\centering\item[*] Note: Significance levels are denoted by * and ** for 0.05 and 0.01, respectively. 
\end{tablenotes}
\label{tab:app:one_source}
\end{table*}

Next, we investigate the effect of training with a subset (ranging from 75\% to 25\%) of the source domain data on the performance of our method. As an example, consider training with 25\% of the source domain data. In this setting, we conducted 25 experiments, each of which randomly sampled 25\% of the data from the Politics and Entertainment domains to train our method. 
Table~\ref{tab:app:partial_source} reports the average performance of our method (with standard deviations in parentheses) in each setting. As expected, the performance of our method declines as the proportion of the source domain data used for training decreases from 75\% to 25\%.
At both the 50\% and 25\% levels, the performance of our method drops significantly across all metrics.

\begin{table*}[htbp]
\centering
\caption{Sensitivity Analysis of Partial Source Domain Data.}
\def\arraystretch{1.3}
\begin{tabular}{ p{4.0cm} p{2.5cm}<{\centering} p{2.5cm}<{\centering} p{2.5cm}<{\centering} p{2.5cm}<{\centering} } 
\toprule 
 & Precision & Recall & F1-score & F2-score \\
\midrule
\multirow{2}{*}{DACA}  & 0.716 & 0.745 & 0.719 & 0.739 \\ \addlinespace[-1ex]
& (0.030) & (0.049) & (0.034) & (0.043) \\ 
\multirow{2}{*}{DACA (75\% of Source Data)}  & 0.708 & 0.724$^{*}$ & 0.704 & 0.721 \\ \addlinespace[-1ex]
& (0.039) & (0.033) & (0.038) & (0.039) \\
\multirow{2}{*}{DACA (50\% of Source Data)}  & 0.681$^{**}$ & 0.700$^{**}$ & 0.684$^{**}$ & 0.696$^{**}$ \\ \addlinespace[-1ex]
& (0.030) & (0.041) & (0.035) & (0.037) \\
\multirow{2}{*}{DACA (25\% of Source Data)}  & 0.669$^{**}$ & 0.681$^{**}$ & 0.665$^{**}$ & 0.679$^{**}$ \\ \addlinespace[-1ex]
& (0.044) & (0.031) & (0.038) & (0.031) \\
\bottomrule 
\end{tabular}%
\begin{tablenotes}
\centering\item[*] Note: Significance levels are denoted by * and ** for 0.05 and 0.01, respectively. 
\end{tablenotes}
\label{tab:app:partial_source}
\end{table*}

Finally, we examine the impact of using partially labeled source domain data on model performance and report the evaluation results in Table~\ref{tab:app:partial_source_labeled}.
In this case, we used all the source domain data for training, but only a portion of it (e.g., 75\%) was labeled. For example, in an experiment, we randomly sampled 25\% of the source domain instances and removed their labels, resulting in 75\% of the instances being labeled.
Among the three modules of our method, the classification and concept alignment modules require labeled data as input.
Since labeled source domain data provides valuable signals for training these two modules, the performance of our method degrades as the percentage of labeled source domain data decreases from 75\% to 25\%.

\begin{table*}[htbp]
\centering
\caption{Sensitivity Analysis of Partially Labeled Source Domain Data.}
\def\arraystretch{1.3}
\begin{tabular}{ p{5cm} p{2.5cm}<{\centering} p{2.5cm}<{\centering} p{2.5cm}<{\centering} p{2.5cm}<{\centering} } 
\toprule 
& Precision & Recall & F1-score & F2-score \\
\midrule
\multirow{2}{*}{DACA}  & 0.716 & 0.745 & 0.719 & 0.739 \\ \addlinespace[-1ex]
& (0.030) & (0.049) & (0.034) & (0.043) \\ 
\multirow{2}{*}{DACA (75\% Labeled Source Data)}  & 0.696 & 0.720$^{*}$ & 0.695$^{*}$ & 0.715$^{*}$ \\ \addlinespace[-1ex]
& (0.048) & (0.037) & (0.034) & (0.034) \\
\multirow{2}{*}{DACA (50\% Labeled Source Data)}  & 0.678$^{**}$ & 0.699$^{**}$ & 0.681$^{**}$ & 0.695$^{**}$ \\ \addlinespace[-1ex]
& (0.049) & (0.038) &  (0.031) & (0.035) \\
\multirow{2}{*}{DACA (25\% Labeled Source Data)}  & 0.662$^{**}$ & 0.681$^{**}$ & 0.654$^{**}$ &  0.677$^{**}$ \\ \addlinespace[-1ex]
& (0.036) & (0.035) & (0.035) & (0.040) \\
\bottomrule 
\end{tabular}%
\begin{tablenotes}
\centering\item[*] Note: Significance levels are denoted by * and ** for 0.05 and 0.01, respectively. 
\end{tablenotes}
\label{tab:app:partial_source_labeled}
\end{table*}

\subsection{Heterogeneity in Source-Target Domain Closeness for Domain Adaption}
\label{appendix:heterogeneity_info}

We investigate the impact of the closeness between source and target domains on the performance of our method. To this end, we conducted additional sensitivity analyses using a public dataset of Chinese news \citepec{nan2021mdfend2}, which spans a broad range of domains. 
Specifically, we employed news data in five different domains, as summarized in Table \ref{tab:app:datastat_cn_r3}. 
Using labeled news from a source domain listed in Table \ref{tab:app:datastat_cn_r3}, we applied our DACA method to predict the label for each piece of unlabeled news in the target COVID domain. 
Table \ref{tab:app:cn_one_source} presents the average precision, recall, F1-score, and F2-score (with standard deviations in parentheses) for each source domain across 25 experiments. For example, when using labeled news from the healthcare domain as the source to predict labels of unlabeled news in the target COVID domain, our method achieves a precision of 0.824, a recall of 0.834, an F1-score of 0.822, and an F2-score of 0.832.

\begin{table*}[htbp]
\centering
\caption{Summary Statistics of Chinese News Dataset (Five Domains). }
\def\arraystretch{1.3}\begin{tabular}{lccc}
\toprule 
Domain & Number of true news & Number of fake news \\
\midrule
COVID (Infodemic / Target) & 201 & 124  \\
Healthcare (Source) & 284 & 391  \\ 
Politics (Source) & 306 & 546 \\
Entertainment (Source) & 1000 & 440   \\ 
Military (Source) & 121 & 222 \\ 
\bottomrule 
\end{tabular}%
\label{tab:app:datastat_cn_r3}
\end{table*}%

The experimental results in Table \ref{tab:app:cn_one_source} suggest that concept alignment and domain adaptation are more effective when source and target domains are more closely related. As shown, the performance of our method increases as we change the source domain from military (least related to the COVID domain) to healthcare (most related to the COVID domain).\footnotemark\footnotetext{To compute the distance between the target COVID domain and each source domain, we first embedded each Chinese news article in each domain as a numerical vector using BERT-wwm \citepec{cui2021pre2}). Next, we calculated the distance between domains using the single-kernel Maximum Mean Discrepancy (MMD) \citepec{borgwardt2006integrating2}. Specifically, the MMD distances between the COVID domain and the healthcare, politics, entertainment, and military domains are 0.071, 0.113, 0.178, and 0.245, respectively.} 
In addition, the experimental results also indicate that a distant source domain (e.g., military in the experiments) still provides signals for classifying completely unlabeled news in the target COVID domain.

\begin{table*}[ht]
\centering
\caption{Performance of Our Method Across Different Source Domains.}
\def\arraystretch{1.3}
\begin{tabular}{ p{3.5cm} p{2.5cm}<{\centering} p{2.5cm}<{\centering} p{2.5cm}<{\centering} p{2.5cm}<{\centering} } 
\toprule 
Source Domain & Precision & Recall & F1-score & F2-score \\
\midrule
\multirow{2}{*}{Healthcare}  & 0.824 & 0.834 & 0.822 & 0.832 \\ \addlinespace[-1ex]
& (0.016) & (0.012) & (0.014) & (0.013) \\ 
\multirow{2}{*}{Politics}  & 0.750 & 0.724 & 0.719 & 0.729 \\ \addlinespace[-1ex]
& (0.047) & (0.036) & (0.038) & (0.038) \\
\multirow{2}{*}{Entertainment}  & 0.698 & 0.702 & 0.700 & 0.701\\ \addlinespace[-1ex]
& (0.059) & (0.047) & (0.058) & (0.055) \\
\multirow{2}{*}{Military}  & 0.699 & 0.687 & 0.666 & 0.689 \\ \addlinespace[-1ex]
& (0.037) & (0.055) & (0.042) & (0.060) \\
\bottomrule 
\end{tabular}%
\label{tab:app:cn_one_source}
\end{table*}

Labeled information (true or false information) in a source domain provides both signals and noise for classifying unlabeled information in the target domain. Signals arise because misinformation shares common characteristics across domains, such as emotional news headlines and inconsistencies between news headlines and news content \citepec{zhou2020fake_ec}. However, according to Theorem~\ref{theorem:single}, covariate and concept shifts between source and target domains introduce noise. Our method leverages signals in a source domain and mitigates noise in the domain to classify unlabeled information in the target domain. Therefore, as reported in Table \ref{tab:app:cn_one_source}, a distant source domain (i.e., military) is still useful for classifying information in the target domain. Moreover, a source domain that is more closely related to the target domain incurs smaller covariate and concept shifts between them. As a result, our method is more effective in classifying unlabeled target domain information when applied to a source domain that is more closely related to the target domain. 

We also evaluated the performance of two benchmarks under different source-target domain pairs: CANMD \citepec{yue2022contrastive2} and DANN \citepec{li2021multi2}. The performance of these methods decreases as the source domain becomes more distant from the target COVID domain. For example, the recall of CANMD drops from 0.788 to 0.647 as the source domain changes from healthcare to military. The performance variations of these methods are consistent with those of our method. The evaluation results further demonstrate that a domain adaptation method is more effective when source and target domains it employs are more closely related. 

\subsection{Performance Evaluation under a Chronological Setting }
\label{appendix:chro}

We evaluated the performance of our method from a chronological perspective. As a first step, we attempted to retrieve the publication date for each news article in the COVID domain. 
To this end, we accessed the raw data through a mirror version provided by Opendatalab\footnotemark\footnotetext{See \url{https://github.com/opendatalab}. We used the mirror version because the original Google Drive link (\url{https://drive.google.com/drive/folders/1gd4AvT6BxPRtymmNd9Z7ukyaVhae5s7U?usp=sharing}) associated with the dataset has expired.}. 
We then extracted the publication date of a news article from its corresponding source-page URL. For example, the publication date of the article with the URL \url{https://www.snopes.com/news/2020/03/19/stopping-coronavirus-what-does-the-evidence-say-is-best/} can be directly identified as 2020-03-19. 
Using this approach, we successfully identified the publication dates of 2,703 articles with valid URLs containing date information. Among these articles, the earliest publication date is 2020-01-01, and the latest is 2020-06-30. 
We then partitioned these articles into two stages: Stage 1, which contains 755 articles published between 2020-01-01 and 2020-03-31, and Stage 2, which contains 1,948 articles published between 2020-04-01 and 2020-06-30. We observed a higher proportion of fake news in Stage 1 (43\%) than in Stage 2 (31\%).

\begin{table*}[htbp]
\centering
\caption{Performance Comparison between DACA and MK-MMD across Chronological Stages.}
\def\arraystretch{1.2}
\begin{tabular}{ p{3.5cm} p{2cm}<{\centering} p{2cm}<{\centering} p{2cm}<{\centering} p{2cm}<{\centering} } 
\toprule 
Method & Precision & Recall & F1-score & F2-score \\
\midrule
\multirow{2}{*}{DACA(Stage 1)} & \textbf{0.670} & \textbf{0.663} & \textbf{0.667} & \textbf{0.665} \\ \addlinespace[-1ex]
& (0.069) & (0.074) & (0.075) & (0.071) \\ 
\multirow{2}{*}{MK-MMD(Stage 1)} & 0.625$^{**}$ & 0.623$^{**}$ & 0.624$^{**}$ & 0.624$^{**}$ \\ \addlinespace[-1ex]
& (0.072) & (0.064) & (0.062) & (0.068) \\

\midrule

\multirow{2}{*}{DACA(Stage 2)} & \textbf{0.721} & \textbf{0.716} & \textbf{0.717} & \textbf{0.717} \\ \addlinespace[-1ex]
& (0.057) & (0.074) & (0.072) & (0.060) \\ 
\multirow{2}{*}{MK-MMD(Stage 2)} & 0.666$^{**}$ & 0.671$^{**}$ & 0.668$^{**}$ & 0.670$^{**}$ \\ \addlinespace[-1ex]
& (0.052) & (0.054) & (0.052) & (0.052) \\ 

\bottomrule 
\end{tabular}%
\begin{tablenotes}
\centering\item[*] Note: Significance levels are denoted by * and ** for 0.05 and 0.01, respectively. 
\end{tablenotes}
\label{tab:time}
\end{table*}

We conducted additional experiments to compare the performance of our DACA method with that of the best-performing benchmark (MK-MMD) under chronological and cumulative settings. Specifically, for Stage 1, we trained each method using COVID news articles published up to 2020-03-31, together with source-domain news articles, and evaluated its performance on Stage 1 COVID news articles. For Stage 2, we trained each method using COVID news articles published up to 2020-06-30, along with source-domain news articles, and evaluated its performance on Stage 2 COVID news articles.
Table \ref{tab:time} reports the average precision, recall, F1-score, and F2-score of DACA and MK-MMD in these two stages. As reported, our method significantly outperforms MK-MMD across all metrics. 
Specifically, our method improves precision, recall, F1-score, and F2-score by 7.16\%, 6.40\%, 6.77\%, and 6.55\%, respectively, in Stage 1, and by 8.22\%, 6.76\%, 7.33\%, and 6.96\%, respectively, in Stage 2.

\bibliographystyleec{informs2014} 
\bibliographyec{reference_ec} 

\end{appendices}

\end{document}